\documentclass[final]{cvpr}
\usepackage{url}
\usepackage{graphicx}
\usepackage{tabularx}
\usepackage{comment}
\usepackage{makecell}
\usepackage{pifont}
\usepackage{multirow}
\usepackage{subfigure}
\usepackage[misc]{ifsym}
\usepackage{afterpage}
\usepackage{setspace}
\usepackage{breakcites}
\usepackage{seqsplit}

\usepackage{times}
\usepackage{epsfig}
\usepackage{graphicx}
\usepackage{amsmath}
\usepackage{amssymb}
\usepackage{wrapfig}
\usepackage{hyphenat}
\usepackage{soul,color}
\usepackage{amsopn}
\usepackage{amsmath}
\usepackage{amssymb}
\usepackage{tabularx}
\usepackage{subfigure}

\definecolor{Red}{cmyk}{0,1,1,0}
\definecolor{Green}{cmyk}{1,0,1,0}
\definecolor{Cyan}{cmyk}{1,0,0,0}
\definecolor{Purple}{cmyk}{0.45,0.86,0,0}
\definecolor{Rosolic}{cmyk}{0.00,1.00,0.50,0}
\definecolor{Blue}{cmyk}{1.00,1.00,0.00,0}
\definecolor{BlueViolet}{cmyk}{0.86,0.91,0,0.04}
\definecolor{NavyBlue}{cmyk}{0.94,0.54,0,0}

\newcommand{\myparagraph}[1]{\vspace{0.1em}\noindent\textbf{#1}}
\usepackage[pagebackref=true,breaklinks=true,colorlinks,bookmarks=false]{hyperref}

\def\cvprPaperID{} % *** Enter the CVPR Paper ID here
\def\confYear{CVPR 2021}

\begin{document}

%%%%%%%%% TITLE
% temp

\title{NeuralHumanFVV: Real-Time Neural Volumetric Human Performance Rendering using RGB Cameras}

\author{Xin Suo\textsuperscript{1} \;\, Yuheng Jiang\textsuperscript{1} \;\, Pei Lin \textsuperscript{1} \;\, Yingliang Zhang\textsuperscript{2} \;\, Kaiwen Guo \textsuperscript{3} \;\, Minye Wu\textsuperscript{1}\;\, Lan Xu\textsuperscript{1,4}} 

\makeatletter
\let\@oldmaketitle\@maketitle% Store \@maketitle
\renewcommand{\@maketitle}{
	\@oldmaketitle% Update \@maketitle to insert...
	\centering
	\vspace{-8mm}
	{\large \textsuperscript{1}ShanghaiTech University}\quad \quad
	{\large \textsuperscript{2}Degene}\quad \quad
	{\large \textsuperscript{3}Google}\\
	{\large	\textsuperscript{4}Shanghai 
Engineering Research Center of Intelligent Vision and Imaging}%\\
% 	{\large	\textsuperscript{3}University of Chinese Academy of Sciences}\\
% 	{\large	\textsuperscript{4}Shanghai Institute of Microsystem and Information Technology}
	%{\tt\normalsize lxuan@connect.ust.hk \{wxu,golyanik,mhaberma,theobalt\}@mpi-inf.mpg.de fanglu@sz.tsinghua.edu.cn} \\
	\vspace{8mm}
}
\makeatother

\maketitle

%%%%%%%%% ABSTRACT
\begin{abstract}
4D reconstruction and rendering of human activities is critical for immersive VR/AR experience.
% but it suffers from inherent self-scanning constraint and consequent fragile tracking under the monocular setting. 
Recent advances still fail to recover fine geometry and texture results with the level of detail present in the input images from sparse multi-view RGB cameras. 
In this paper, we propose NeuralHumanFVV, a real-time neural human performance capture and rendering system to generate both high-quality geometry and photo-realistic texture of human activities in arbitrary novel views.
We propose a neural geometry generation scheme with a hierarchical sampling strategy for real-time implicit geometry inference, as well as a novel neural blending scheme to generate high resolution (e.g., 1k) and photo-realistic texture results in the novel views.
Furthermore, we adopt neural normal blending to enhance geometry details and formulate our neural geometry and texture rendering into a multi-task learning framework. 
Extensive experiments demonstrate the effectiveness of our approach to achieve high-quality geometry and photo-realistic free view-point reconstruction for challenging human  performances.
%
% Our approach removes the need for multi-view studio settings and enables a consumer-accessible solution for volumetric capture.

\end{abstract}

%%%%%%%%% BODY TEXT

\section{Introduction}
% 1. top view of importance for
% both performance capture and novel view sythesis
The rise of virtual and augmented reality (VR and AR) to present information in an immersive way has increased the demand of the 4D (3D spatial plus 1D time) content generation.
Further reconstructing human activities and providing photo-realistic rendering from a free viewpoint conveniently evolves as a cutting-edge yet bottleneck technique.
% and has recently attracted substantive attention of both the computer vision and computer graphics communities.

\begin{figure}[tbp] 
	\centering 
	\includegraphics[width=1\linewidth]{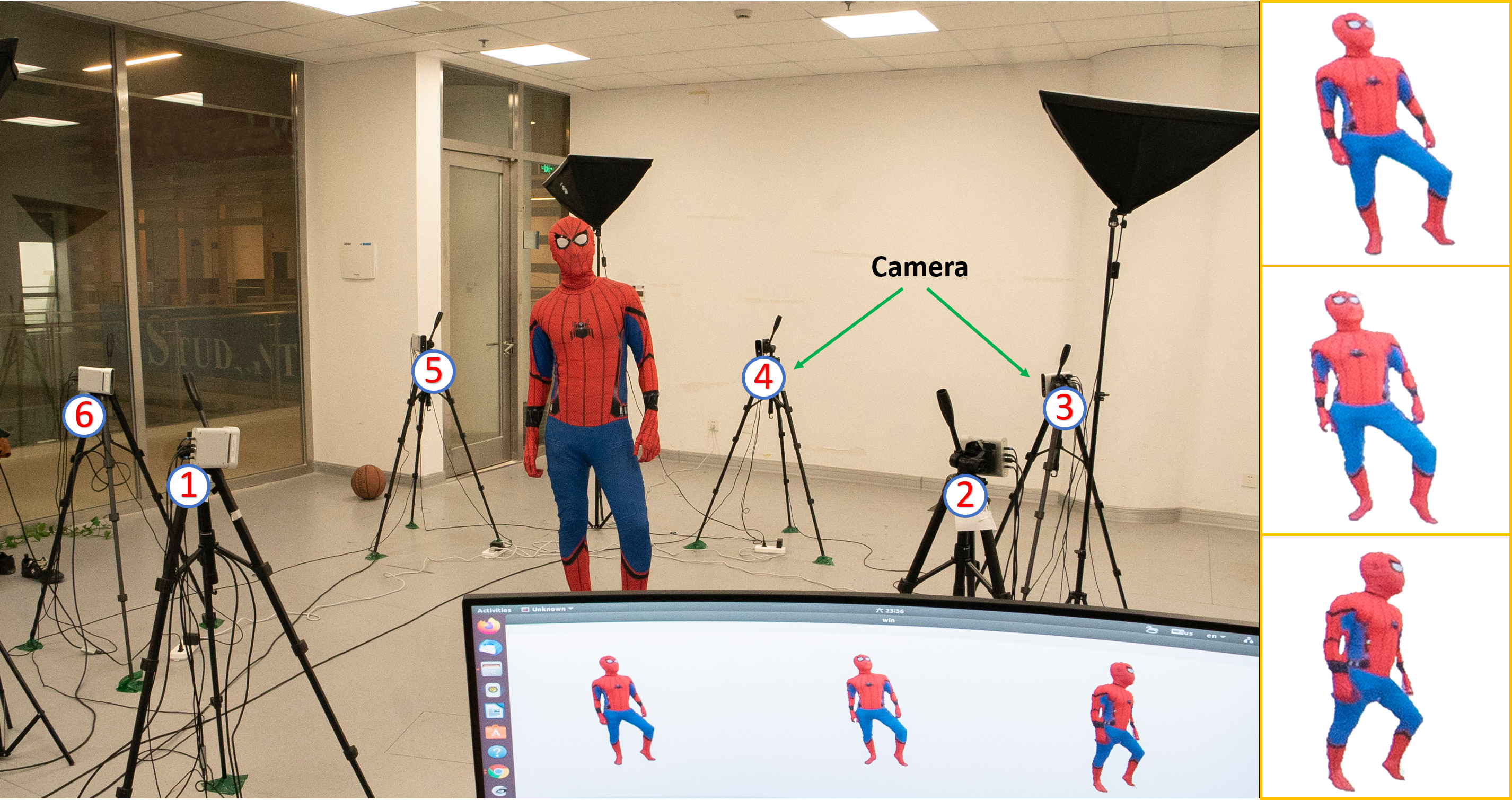} 
	\caption{Our NeuralHumanFVV achieves real-time and photo-realistic reconstruction results of human performance in novel views, using only 6 RGB cameras.} 
	\label{fig:fig_1_teaser} 
	%\vspace{-8pt} 
	\vspace{-10pt} 
\end{figure}

% 2. Issues of existing methods, especially for immersive tele-presence
% 2.1 depth-based methods: 
Early solutions~\cite{li2009robust,HaoliTemplate,Templaterealtime,collet2015high} require pre-scanned templates or two to four orders of magnitude more time than is available for daily usages such as immersive tele-presence.
Recently, volumetric approaches have enabled real-time human performance reconstruction and eliminated the reliance of a pre-scanned template model, by leveraging the RGBD sensors and modern GPUs.
 %leading to the high restriction of the wide applications for daily usage.
The high-end solutions~\cite{dou-siggraph2016,motion2fusion,TotalCapture,UnstructureLan} rely on multi-view studio setup to achieve high-fidelity reconstruction and rendering in a novel view but are expensive and difficult to be deployed, 
while the low-end approaches~\cite{Newcombe2015,KillingFusion2017cvpr,FlyFusion,DoubleFusion,robustfusion} adopt the most handy monocular setup with a temporal fusion pipeline~\cite{KinectFusion} but suffer from inherent self-occlusion constraint.
% to achieve complete reconstruction. Recent state-of-the-art methods further utilize human prior~\cite{BodyFusion,} or additional body-worn inertial devices~\cite{HybridFusion} to improve the tracking accuracy. However, all these single-view approaches suffer from careful and orchestrated motions, especially for a tedious self-scanning process where the performer need to turn around carefully to obtain complete reconstruction. 
% When the captured model is incomplete, the non-rigid tracking in those newly fused regions is fragile, leading to inferior results and impractical usage for VR/AR applications such as immersive tele-presence.
%
Moreover, these approaches above rely on depth cameras which are not as cheap and ubiquitous as color cameras.

% 2.2 data-driven RGB solutions; Neural rendering solution
The recent learning-based techniques enable robust human attribute reconstruction~\cite{Mehta2017,PIFU_2019ICCV,DeepHuman_2019ICCV,MonoPort} using only RGB input.
% PIFU, PIFUHD monoPort: real-time geometry with fine details but only per-vertex color in the novel view
In particular, the approaches PIFu~\cite{PIFU_2019ICCV} and PIFuHD~\cite{PIFuHD} utilize pixel-aligned implicit function to reconstruct clothed humans with fine geometry details, while MonoPort~\cite{MonoPort} further enables real-time inference in a novel view.
However, these methods fail to generate compelling photo-realistic texture due to the reliance of implicit texture representation.
% Neural novel view synthesis: per-scenes training, non-realtime performance
On the other hand, neural rendering techniques~\cite{NeuralVolumes,CtViewControl,Wu_2020_CVPR,nerf,RTH_nvs,FreeViewSynthesis} bring huge potential for photo-realistic novel view synthesis.
However, existing solutions rely on per-scene training or are hard to achieve real-time performance due to the heavy network and the complicated 3D representation.
Moreover, few researchers explore to combine volumetric geometry modeling and photo-realistic novel view synthesis of human performance in a data-driven manner simultaneously, especially under the light-weight multi-RGB and real-time setting.

% Such data-driven visual cues encode various prior information of human models such as motion~\cite{OpenPose,Mehta2017,HMR18} or geometry~\cite{PIFU_2019ICCV,TEX2SHAPE_2019ICCV,DeepHuman_2019ICCV}.
%
% However, researchers pay less attention to strengthen the volumetric performance capture with various data-driven visual cues, especially for template-less setting.

% 3. We propose a robust scheme for human volumetric capture, using only a single RGBD sensor.
% which outperforms existing state-of-the-art approaches significantly.
% Key novelty
In this paper, we attack the above challenges and present \textit{NeuralHumanFVV} -- a real-time human neural volumetric rendering system using only light-weight and sparse RGB cameras surrounding the performer.
As illustrated in Fig.~\ref{fig:fig_1_teaser}, our novel approach generates both high-quality geometry and photo-realistic texture of human activities in arbitrary novel views, whilst still maintaining real-time computation and light-weight setup.

% 4. Our technical Pipeline illustration
Generating such a human free-viewpoint video by combining volumetric geometry modeling and neural texture synthesis in a data-driven manner is non-trivial.
% 4.1 our key idea:
Our key idea is to encode the local fine-detailed geometry and texture information of the adjacent input views into the novel target view, besides utilizing the inherent global information from our multi-view setting.
% 4.2 neural geometry generation: to implicitly reason about the underlying geometry
To this end, we first introduce a neural geometry generation scheme to implicitly reason about the underlying geometry in a novel view.
With a hierarchical sampling strategy along the camera rays in a coarse-to-fine manner, we achieve real-time detailed geometry inference.
% 4.3 neural texture blending
Then, based on the geometry proxy above, a novel neural blending scheme is proposed to map the input adjacent images into a photo-realistic texture output in the target view, through efficient occlusion analysis and blending weight learning.
A boundary-aware upsampling strategy is further adopted to generate high resolution (e.g., 1k) novel view synthesis result without sacrificing the real-time performance.
% 4.4 normal refinement
Finally, we recover the normal information in the target view using the same neural blending strategy, which not only enhances the output fine-grained geometry details but also combines our neural geometry generation and texturing blending into a multi-task learning framework. 
%5. Our technical contribution
To summarize, our main contributions include:
\begin{itemize} 
	\setlength\itemsep{0em}
	\item We present a real-time human performance rendering approach, which is the first to reconstruct high quality geometry and photo-realistic texture results in a novel view using sparse multiple RGB cameras, achieving significant superiority to existing state-of-the-arts.
	
	\item We propose an efficient neural implicit generation scheme to recover fine geometry details in the novel view via a hierarchical and coarse-to-fine strategy.
	%combine data-driven occupancy representation with volumetric fusion only using the front-view input.
	
	\item We propose a novel neural blending scheme to provide high-resolution and photo-realistic texture result as well as normal result to further refine the geometry. 
\end{itemize} 
% 1) A robust volumetric capture pipeline
% 2) DeepScan
% 3) Detector-based tracking

% Our hybrid optimization not only handles challenging fast human motions but also owns the reinitialization ability to recover from tracking failures and the disappear-reoccur scenarios, while the volumetric fusion strategy estimates material-dependent motion tracking behavior to achieve robust and precise geometry update and avoid deteriorated fusion model caused by challenging fast motion and self-occulusion.
%Given the aforementioned distinctiveness, UnstructuredFusion serves as a good compromising settlement between over- demanding hardware setup and high-quality reconstruction, promoting potential applications of 4D reconstruction in immersive telepresence and supporting better immersive/interactive experiences.
%Our new model achieves high quality dense human performance capture results on our new challenging dataset, demonstrating, qualitatively and quantitatively, the advantages of our approach over previous work.

\begin{figure*}[t] 
    \vspace{-4ex}
	\begin{center} 
		\includegraphics[width=\linewidth]{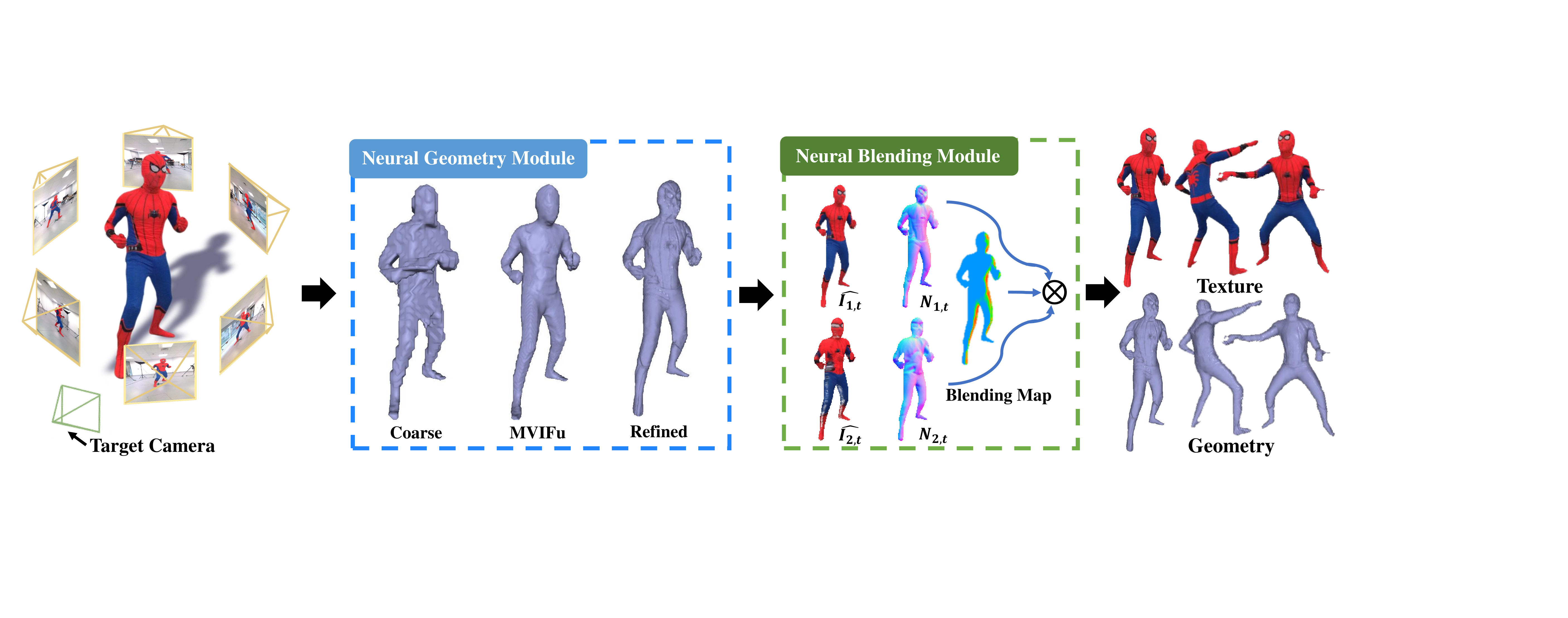} 
	\end{center} 
	\vspace{-2ex}
	\caption{The pipeline of NeuralHumanFVV. Assuming the video input from six RGB cameras surrounding the performer, our approach consists of a neural geometry generation stage (Sec.~\ref{sec:geometry}) and a neural blending stage (Sec.~\ref{sec:texture}) to generate live 4D rendering results.} 
	\label{fig:fig_2_overview} 
	\vspace{-10pt}
\end{figure*}

\section{Related Work} 
\myparagraph{Human Performance Capture.}
Markerless human performance capture~\cite{BreglM1998,TheobASST2010} technologies have been widely investigated to generate human free-viewpoint video or geometry reconstruction.
% high-end multi-view solutions
The high-end approaches require studio-setup with hundreds of cameras and a controlled imaging environment~\cite{StollHGST2011,liu2013markerless,joo2015panoptic,collet2015high,TotalCapture,TheRelightables} to produce high quality surface motion and appearance reconstruction.
% sigle-view, handheld or drone-based
Some recent work only relies on the light-weight and single-view setup~\cite{MonoPerfCap,LiveCap2019tog,EventCap} and even enables hand-held capture~\cite{Wu2013,SMPLX2019,Xiang_2019_CVPR} or drone-based capture~\cite{FlyCap}.
However, these methods require the pre-scanned template or naked human model.
% Volumetric methods
Only recently, monocular free-form dynamic reconstruction methods~\cite{Newcombe2015,guo2017real,DoubleFusion,FlyFusion,robustfusion} with real-time performance have been proposed by combining the volumetric fusion \cite{Curless1996} and the nonrigid tracking~\cite{sumner2007embedded,li2009robust,zollhofer2014real} using RGBD camera.
However, these monocular methods still suffer from the inherent self-occlusion constraint and cannot capture the motions in occluded regions.
The light-weight multi-view solutions~\cite{dou-siggraph2016,motion2fusion,UnstructureLan} serve as a good compromising settlement between over-demanding hardware setup and high-fidelity reconstruction but still rely on 3 to 8 RGBD streams as input.
Comparably, our approach enables real-time high-quality geometry and photo-realistic texture reconstruction in novel views only using 6 RGB cameras surrounding the performer.

\myparagraph{Data-Driven Human Modeling.}
Early human modeling approaches~\cite{Shotton2011,Ganapathi10} formulate the discriminative performance capture into a regression or classification problem using machine learning techniques.
% Learn motions, geometry, garments and texture from RGB input
With the advent of deep neural networks, recent approaches obtain various human attributes successfully from only RGB input.
% 1. shape and pose
Some recent work~\cite{OpenPose,Mehta2017,HMR18,DensePose} learns the skeletal pose and even human shape prior by using human parametric models~\cite{SCAPE2005,SMPL2015}.
% 2. geometry
Various approaches~\cite{DetailDepth_2019ICCV,People3D_2019ICCV,TEX2SHAPE_2019ICCV,DeepHuman_2019ICCV} propose to predict human geometry from a single RGB image by utilizing parametric human model as a basic estimation.
Several work~\cite{DeepVolumetric_2018ECCV,DeepSDF,occupancy2019CVPR,PIFU_2019ICCV,PIFuHD} further reveals the effectiveness of learning the implicit occupancy directly for textured geometry modeling and even real-time inference~\cite{MonoPort}.
%
% 3. texture, garmentsa
%
Besides, researchers~\cite{bhatnagar2019mgn,lazova3dv2019} propose to fetch the garment or texture information of the human model.
However, these data-driven human modeling methods still fail to recover fine geometry and texture results simultaneously with the level of detail present in the RGB inputs.
In contrast, we explore to combine implicit geometry modeling with novel view synthesis in a data driven manner for real-time, high-quality and photo-realistic human performance rendering, achieving significant superiority to previous methods.

\myparagraph{Neural Rendering.}
The recent progress of neural rendering techniques~\cite{NR_survey,CtViewControl,Wu_2020_CVPR,RTH_nvs} brings huge potential for constructing neural scene representations~\cite{DeepVoxels_CVPR2019,NeuralVolumes,SRN_nips2019,nerf} and photo-realistic novel view blending~\cite{NRWird_CVPR19,DeepBlending,Thies2020Image-guided,FreeViewSynthesis}.
% 1. neural scene representation --> per-scene training
For reconstructing neural scenes, various data representations have been explored, such as  point-clouds~\cite{aliev2019neural,Wu_2020_CVPR}, voxels~\cite{DeepVoxels_CVPR2019,NeuralVolumes} or implicit representations~\cite{SRN_nips2019,nerf,liu2020neural}.
However, dedicated per-scene training is required in these methods when applying the representation to a new scene. 
% 2. IBR and blending based --> free-view synthesis
% \cite{DeepBlending,FreeViewSynthesis}
Various methods~\cite{DeepBlending,FreeViewSynthesis} learn the mapping of features from source images to novel target views to avoid per-scene training, while some recent work~\cite{Thies2020Image-guided,DVS_photo} further models the view-dependent effects.
However, these methods rely on heavy networks or complicated 3D proxies which are unsuitable for real-time applications like immersive telepresense.
% 3. continues view control, latent code to represent complicated geometry.
Chen~\emph{et al.}~\cite{CtViewControl} propose to predict the output texture using implicit underlying geometry, which enables continues view generation from monocular image.
Researchers~\cite{RTH_nvs,DodgeABullet_ECCV2018} also utilize such underlying latent geometry for novel view
synthesis of human performance in the encoder-decoder manner. 
However, these approaches suffer from limited representation ability of a single latent code for complex human inferior texture output.
Besides, some recent methods~\cite{SemiPara_2019CVPR,LookinGood} combine the neural rendering techniques to provide more visually pleasant results under the traditional RGBD fusion pipeline~\cite{dou-siggraph2016}.
Comparably, our method is the first to embrace neural blending into the implicit geometry modeling pipeline under the light-weight multi-RGB and real-time setting, which enables photo-realistic texture and geometry reconstruction in novel views.

%  besides utilizing the inherent global information from our multi-view setting.

\section{Overview}
The proposed NeuralHumanFVV marries implicit volumetric modeling with neural texture rendering, which generates high-quality geometry and photo-realistic texture of human activities in arbitrary novel views in real-time, and enables various applications like immersive telepresense.
Fig.~\ref{fig:fig_2_overview} illustrates the high-level components of our system, which takes 6 RGB videos surrounding the performer as input and generates high-quality novel-view synthesis results in challenging scenarios with various poses, clothing types and topology changes as output.
%A brief introduction of the components of our pipeline is provided as follows.

\myparagraph{Neural Geometry Generation.}
We first utilize the inherent geometry prior from our multi-view setting via the shape-from-silhouette~\cite{sfs} technique.
Then, we adopt the pixel-aligned implicit function~\cite{DeepVolumetric_2018ECCV,PIFU_2019ICCV,PIFuHD} to maintain the complete and continues geometry of the scene.
Differently, we further recover the underlying geometry in novel views with a multi-stage hierarchical sampling strategy along the camera rays which enables both real-time detailed geometry inference and the following neural blending stage (Sec.~\ref{sec:geometry}).

\myparagraph{Neural Blending.} 
The core of our pipeline is to encode the local fine-detailed geometry and texture information of the adjacent input views into the novel target view.
A novel neural blending scheme is proposed to map the input adjacent images into a photo-realistic texture output in the target view, through efficient occlusion analysis and blending weight learning.
A boundary-aware upsampling strategy is further adopted to generate high resolution (e.g., 1k) novel view synthesis result without sacrificing the real-time performance.
We also recover the normal information in the target view using the same neural blending strategy, which not only enhances the output fine-grained geometry details but also formulates our neural geometry and texture generation in a multi-task learning framework (Sec.~\ref{sec:texture}).

\section{NeuralHumanFVV Method}\label{sec:algorithm} 
\subsection{Neural Geometry Reconstruction} \label{sec:geometry}

%We first utilize the inherent geometry prior from our multi-view setting via the shape-from-silhouette~\cite{sfs} technique.
%Then, we adopt the pixel-aligned implicit function~\cite{DeepVolumetric_2018ECCV,PIFU_2019ICCV,PIFuHD} to maintain the complete and continues geometry of the scene.
%Differently, we further recover the underlying geometry in novel views with a multi-stage hierarchical sampling strategy along the camera rays which enables both real-time detailed geometry inference and the following neural blending stage.

\begin{figure}[t] 
	\begin{center} 
		\includegraphics[width=1\linewidth]{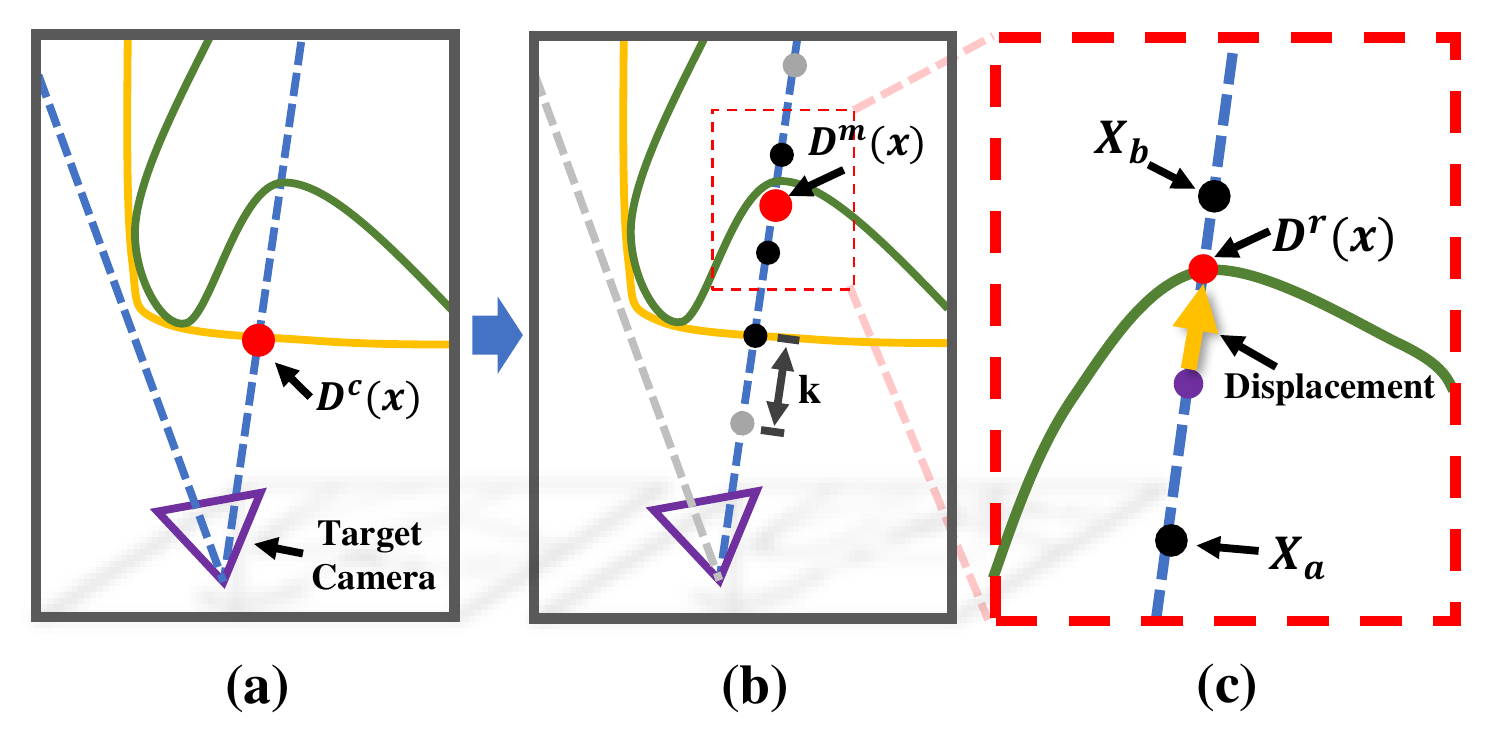}
	\end{center} 
	\vspace{-3mm}
	\caption{Illustration of our hierarchical and coarse-to-fine strategy in neural geometry generation. Orange curves are the coarse geometry surface recovered by SfS; Green curves are the real geometry. Gray dot line and points are discarded in our hierarchical sampling algorithm. (a) is the result of coarse reconstruction; (b) is the result of MVIFu; (c) is the result after refinement. } 
	\label{fig:fig_3_geometry} 
	\vspace{-10pt}
\end{figure}

Given the six RGB images input at each frame, we introduce a coarse-to-fine multi-stage neural geometry reconstruction scheme to generate the inherent detailed human geometry in novel views in real-time, as illustrated in Fig.~\ref{fig:fig_3_geometry}.

\noindent{\bf Coarse Geometry Generation.} Firstly, we extract the coarse inherent geometry prior from our multi-view setting. 
We apply the Shape-from-Silhouette (SfS)~\cite{sfs} algorithm on the human masks segmented off-the-shelf video segmentation method to obtain a coarse human shape. 

\noindent{\bf Accelerated Multi-View Implicit Function.} We extend the pixel-aligned implicit function~\cite{PIFU_2019ICCV,PIFuHD} to our multi-view setting. 
Such multi-view implicit function (MVIFu) maintains the complete and continues geometry of the captured scene, and encodes the human shape priors. 
Similar to \cite{PIFU_2019ICCV}, the implicit function $f$ defines the occupancy of every 3D point $X$ in the space, which is formulated as:
\begin{equation}
\begin{split}
f(\phi(X),z(X)) &= s : s\in[0.0, 1.0], \\
\phi(X) &= \frac{1}{n}\sum_i^n F_i(\pi_i(X)),
\end{split}
\end{equation}
where $\pi_i()$ projects a 3D point into $i$-th source view; $z(X)$ is the depth value in the camera coordinate space. 
The projected image feature at the pixel coordinate $x$ is formulated as $F_i(x)=g(I_i(x))$, where $g$ denotes a feature extraction network. 

Since extracting the whole human geometry is expensive and unnecessary for real-time immersive application, the MVIFu in our pipeline only generates geometry explicitly in the novel view.
Thus, we sample evenly spaced 3D points from near to far based on the coarse geometry along each pixel ray with a distance of $k$ in the target view.
We select first two adjacent sample points to define the range where the depth value of the ray falls.
Specifically, let $X_a$ and $X_b$ be these two points, and $s_a, s_b$ are their occupancy, which satisfy $z(X_a)< z(X_b)$ and $s_a<0.5, s_b \geq 0.5$. 
The predicted depth of this pixel $x$ is given by $D^m(x) = \frac{z(X_a)+z(X_b)}{2}$. 
Moreover, point sampling after $X_b$ can be early terminated. 
We also prune unnecessary sample points on the background pixel rays outside the coarse geometry generated by SfS algorithm, which inherently contains the whole performer so as to enable real-time reconstruction.

\noindent{\bf Depth Fine-tuning.} 
%MVFu provides a depth range and an approximated depth for each pixel of human. 
%Even though the MVIFu result is much more accurate than the coarse geometry, 
The geometry $D^m$ obtained through our accelerated MVIFu is still over smooth because of the depth averaging.
In order to recover the geometry details (e.g. clothes wrinkles), we introduce a hierarchical sampling strategy. 
Specifically, we introduce a depth fine-tuning network $h$ which takes the feature of the midpoint between two selected sample points as input, and outputs the displacement of depth value:
\begin{equation}
h(\phi(\frac{X_a+X_b}{2})) = o : o\in[-1.0, 1.0].
\end{equation}
Here, positions on this segment are mapped from $-1.0$ to $1.0$ linearly, and the refined depth value $D^r(x)$ can be composed with the offset $o$ to encode more geometry details:
\begin{equation}
\begin{split}
D^r(x) = D^m(x) + k\cdot\frac{o+1}{2}.
\end{split}
\end{equation}

%--------------------------------------------------------------------------------------------------%
%--------------------------------------------------------------------------------------------------%
%--------------------------------------------------------------------------------------------------%

\begin{figure*}[h] 
    \vspace{-15pt} 
	\begin{center} 
		\includegraphics[width=1\linewidth]{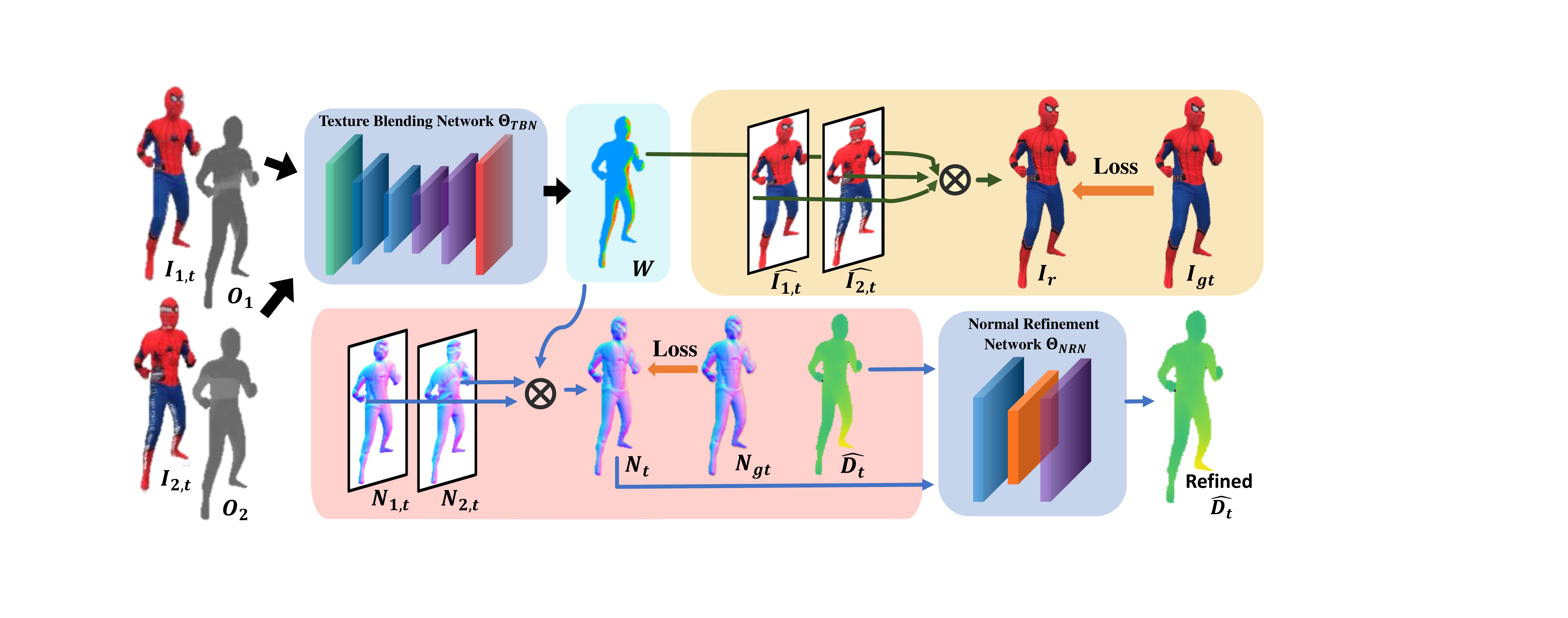} 	
	\end{center} 
	\vspace{-2ex}
	\caption{Illustration of our neural blending scheme, which encodes the local fine-detailed geometry and texture information of the adjacent input views into the novel target view.} 
	\label{fig:fig_3_blending} 
	\vspace{-10pt} 
\end{figure*}

%The core of our pipeline is to encode the local fine-detailed geometry and texture information of the adjacent input views into the novel target view.
%
%A novel neural blending scheme is proposed to map the input adjacent images into a photo-realistic texture output in the target view, through efficient occlusion analysis and blending weight learning.
%
%A boundary-aware upsampling strategy is further adopted to generate high resolution (e.g., 1k) novel view synthesis result without sacrificing the real-time performance

%Input images adjacent to the target camera contain almost all appearance information in the novel view. 
%Existing image-based rendering approaches linearly blend textures from adjacent input views based on geometry proxy. However, their rendering result quality is limited by the geometry proxy. 
\subsection{Neural Blending} \label{sec:texture}
We introduce a neural blending pipeline to encode more local fine-detailed geometry and texture information of the adjacent input views than traditional image-based rendering approaches, so as to produce photo-realistic output in the target view in a data-driven manner, as illustrated in Fig.~\ref{fig:fig_3_blending}.

\myparagraph{Image Warping and Occlusion Analysis.}
Most of the texture information in a target view can be recovered by its only two adjacent input views in our multi-view setting.
Based on this finding, we first generate the depth maps of the target view ($D^r_t$) and the two input views ($D^r_1$ and $D^r_2$, respectively) as described in Sec.~\ref{sec:geometry}.
Then, we use $D^r_t$ to warp the input image $I_1$ and $I_2$ into the target view, denoted by $I_{1,t}$ and $I_{2,t}$.
We also warp source view depth maps into target view and obtain $D^r_{1,t}$ and $D^r_{2,t}$ so as to obtain the 
occlusion map $O_i = D^r_{i,t} - D^r_t (i = 1,2)$, which implies the occlusion information.

\noindent{\bf Texture Blending Network(TBN).} 
$I_{1,t}$ and $I_{2,t}$ may be incorrect due to self-occlusion and inaccurate geometry proxy. 
Simply blending them will raise strong artifacts. 
Thus, we introduce a blending network $\Theta_{TBN}$, which utilizes the inherent global information from our multi-view setting, and fuse local fine-detailed geometry and texture information of the adjacent input views with the pixel-wise blending map $W$, which can be formulated as:
\begin{equation}
\begin{split}
W = \Theta_{TBN}(I_{1,t}, O_1, I_{2,t}, O_2).\\
\end{split}
\end{equation}

\noindent{\bf Boundary-Aware Depth Upsampling.} 
For real-time perfomance, depth maps are generated at low resolution (256$\times$256). 
Aiming to photo-realistic rendering, we need to upsample both the depth map and blending map to 1K resolution.
However, na\"{i}ve upsampling will cause severe zigzag effect near the boundary due to depth inference ambiguity.
Thus, we propose a boundary-aware scheme to refine the human boundary area on the depth map. 
Specifically, we use bilinear interpolation to upsample $D^r_t$. 
Then a erosion operation is applied to extract boundary area. Depth values inside boundary area are recalculated by using the pipeline as described in Sec.~\ref{sec:geometry} and form $\hat{D^r_t}$ at 1K resolution. 
Then we warp the original high resolution input images into the target view with $\hat{D^r_t}$ to obtain $\hat{I_{i,t}}$. 
To this end, our final texture blending result is formulated as:
\begin{equation}
\begin{split}
I_r = \hat{W}\cdot\hat{I_{1,t}} + (1.0-\hat{W})\cdot\hat{I_{2,t}}, \\
\end{split}
\end{equation}
where $\hat{W}$ is the high resolution blending map upsampled by bilinear interpolation directly.

%\subsection{Neural Normal Refinement} \label{sec:normal} 
\noindent{\bf Neural Normal Refinement.}
We apply networks introduced in \cite{PIFuHD} on the input RGB images to inference its normal maps. 
Then, the normal information in the target view is restored via the same neural blending strategy. 
The blended normal map $N_t$ can further enable the geometry refinement. Specifically, we introduce a normal refinement network $\Theta_{NRN}$ to infer the displacement of the target depth map from $N_t$ and $\hat{D^r_t}$, as illustrated in Fig.~\ref{fig:fig_3_blending}.
%Finally, we recover the normal information in the target view using the same neural blending strategy, which not only enhances the output fine-grained geometry details but also combines our neural geometry generation and texturing blending into a multi-task learning framework.

%\begin{equation}
%\begin{split}
%$\widetilde{D_r^t} = \widehat{D_r^t}  + S$ \\
%\end{split}
%\end{equation}

\begin{figure*}[htbp] 
	\begin{center} 
		\includegraphics[width=1\linewidth]{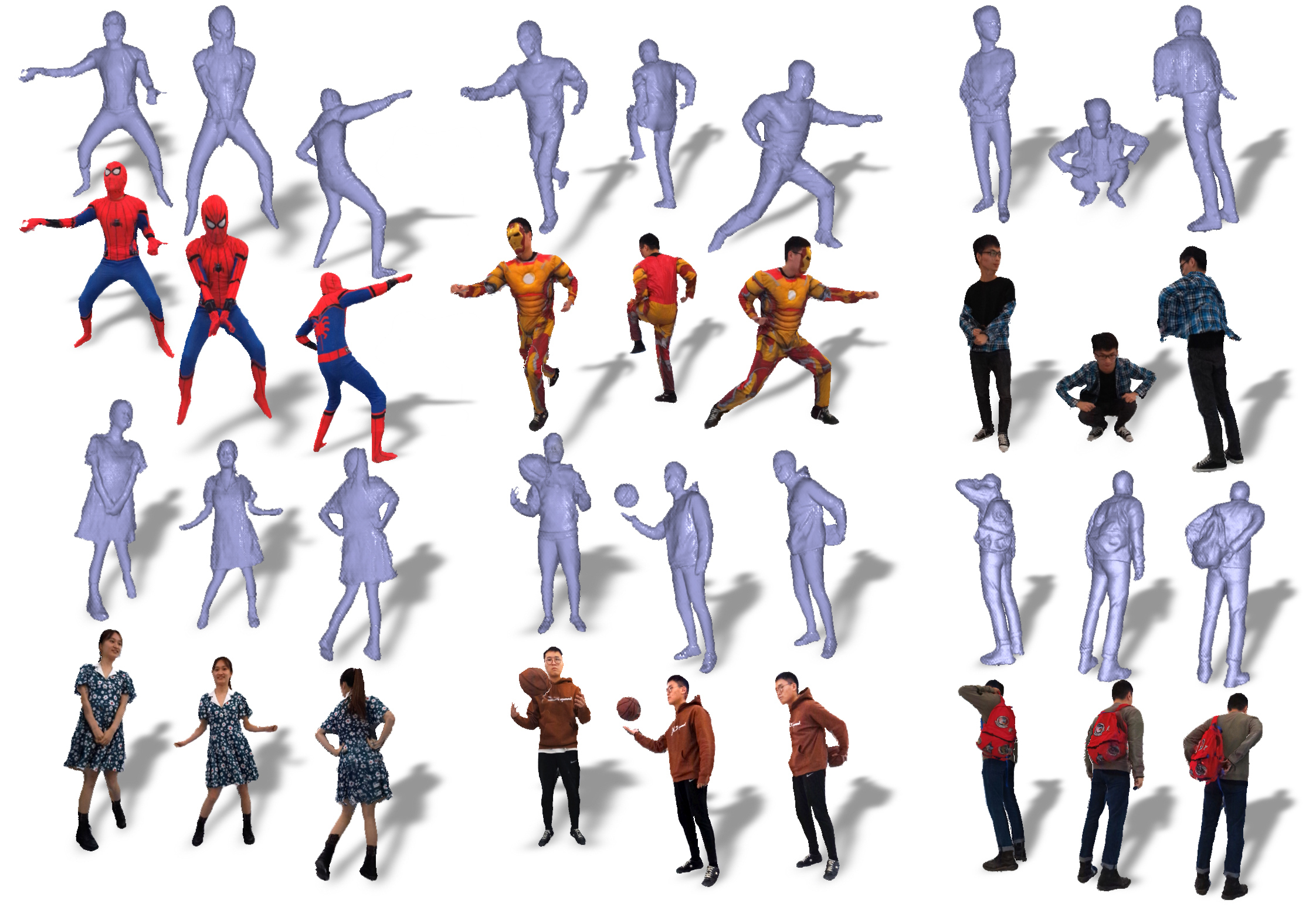} 
	\end{center} 
	\vspace{-10pt}
	\caption{The geometry and texture results of our NeuralHumanFVV on several sequences, including ``spiderman'', ``ironman'', ``undressing'', ``floral dress'', ``basketball'' and ``backpack'' from the upper left to lower right.} 
	\label{fig:fig_all}
	\vspace{-10pt}
\end{figure*}

%--------------------------------------------------------------------------------------------------%
%--------------------------------------------------------------------------------------------------%
%--------------------------------------------------------------------------------------------------%
\subsection{Data and implementation Details} \label{sec:detail} 
The key of our NeuralHumanFVV is to train the neural networks in Sec.~\ref{sec:geometry} and Sec.\ref{sec:texture} properly, including the feature extractor $g$, continuous implicit function $f$, depth fine-tuning network $h$, as well as our TBN $\Theta_{TBN}$ and NRN $\Theta_{NRN}$.
Specifically, $g$ is a U-Net~\cite{ronneberger2015u} and outputs a 64 channels feature maps. 
$f$ represented by MLPs has the same structure in~\cite{PIFU_2019ICCV} but the network dimension is further reduced for real-time performance, while $h$ has the same network architecture, only with a different activation function of the last layer replaced by hyperbolic tangent.
Besides, both our TBN $\Theta_{TBN}$ and NRN $\Theta_{NRN}$ adopt the U-Net structure.

% 2. Dataset setting
We utilize 1820 scans from Twindom~\cite{twindom} and augment the dataset by rigging the 3D model to add more challenging poses, so as to enhance the generation ability of our networks.
We fix the six input camera views as a rig surrounding the performer, and sample 180 virtual target views on a sphere.
Note that all the 3D models locate on the central regions of the sphere and all the cameras face towards the model. 
Our training dataset contains the RGB images, normal maps and depth maps for all the views and models.

% 3.loss setting and training process: g, f,h
For the training of $g$ and $h$ in our MVIFu module, we only use the six input camera views in our dataset, and follow the training procedure similar to previous work~\cite{PIFU_2019ICCV}
Then, we train our depth fine-tuning network $h$ using the corresponding pair-wised data provided by the MVIFu.

% 4.loss setting and training process: TBN and NRN
For the training of our texture blending network $\Theta_{TBN}$, we set out to apply a multi-task learning scheme so as to enable more robust blending weight learning.
The training objective is to make both the blended texture and normal map as close as possible to the ground truth, as these two tasks share the same blending map in our $\Theta_{TBN}$. 
To this end, the loss function includes a appearance term and a normal term with perceptual loss:

\begin{equation}
\begin{split}
\mathcal{L}_{rgb} &= \frac{1}{n} \sum^n_j(\| I^j_r - I^j_{gt}\|^2_2) + \| \varphi(I^j_r) - \varphi(I^j_{gt})\|^2_2), \\
\mathcal{L}_{norm} &= \frac{1}{n} \sum^n_j(\|N^j_t - N^j_{gt}\|^2_2) + \| \varphi(N^j_t) - \varphi(N^j_{gt})\|^2_2), \\
\mathcal{L} &= \lambda\cdot\mathcal{L}_{rgb} + (1.0-\lambda)\cdot\mathcal{L}_{norm},
\end{split}
\end{equation}
where $I_{gt}$ and $N_{gt}$ are the ground truth RGB images and normal maps; $\varphi(\cdot)$ denotes the output features of the third-layer of pretrained VGG-19.

Our normal refinement network (NRN) $\Theta_{NRN}$ need to be adapted to real data, which means we cannot supervise the network training using the same synthetic dataset. 
Thus, we introduce a self-supervise learning scheme where all the training inputs are collected from the real data generated in our pipeline. 
%After apply the displacement map from the network output on the depth map, we can calculate the normal map $\widetilde{N}$ from refined depth. Then let the $\widetilde{N}$ close to the input normal. 
The objective is to minimize the loss function:    
\begin{equation}
\mathcal{L} = \frac{1}{n} \sum^n_j\| \nabla(\hat{D^{r,j}_t}  + \Theta_{NRN}(N^j_t,\hat{D^{r,j}_t})) - N^j_t \|^2_2 
\end{equation}
where $\nabla(\cdot)$ is the operator which calculates the normal map from input depth map.

\section{Experimental Results} 
In this section, we evaluate our NeuralHumanFVV method on a variety of challenging scenarios. 
We run our experiments on a PC with 3.7 GHz Intel i7-8700k CPU $32$GB RAM, and Nvidia GeForce RTX3090 GPU. 
With the live stream data from six RGB cameras, our system generates high-quality geometry and texture results in novel views at 12 fps to enable various interactive immersive applications. 
The whole pipeline costs approximate 80 ms per frame, where the neural geometry generation takes 64 ms and 16 ms for the neural blending stage.
%In all experiments, we use the following empirically determined parameters:  $\lambda_{3D} = 1$, $\lambda_{2D} = 200$, $\lambda_{adj} = 50$, $\lambda_{temp} = 80$,  $\lambda_{sil} = 1.0$, $\lambda_{stab} = 5.0$, and $\lambda_{dist} = 4.0$. For evaluation, we produce a testing dataset where we use different models in the same way as the training dataset.
%
Fig.~\ref{fig:fig_all} demonstrates several results of our NeuralHumanFVV, which can generate free-view high quality geometry and texture results simultaneously.
Noted that our approach can handle human object interaction scenarios with topology changes, such as playing basketball, carrying bag and removing clothes.

%frames of our EventCap results on the proposed dataset. 
%We can see in Fig.~\ref{fig:fig_all} that our results can be precisely overlaid on the latent images (c-d), and that our reconstructed poses are plausible in 3D (e-f).
%We can see that our method can accurately capture the high-frequency temporal motion details, which cannot be achieved by using standard low \textit{fps} videos.

\begin{figure*}[htbp] 
    \vspace{-10pt}
	\begin{center} 
		\includegraphics[width=0.9\linewidth]{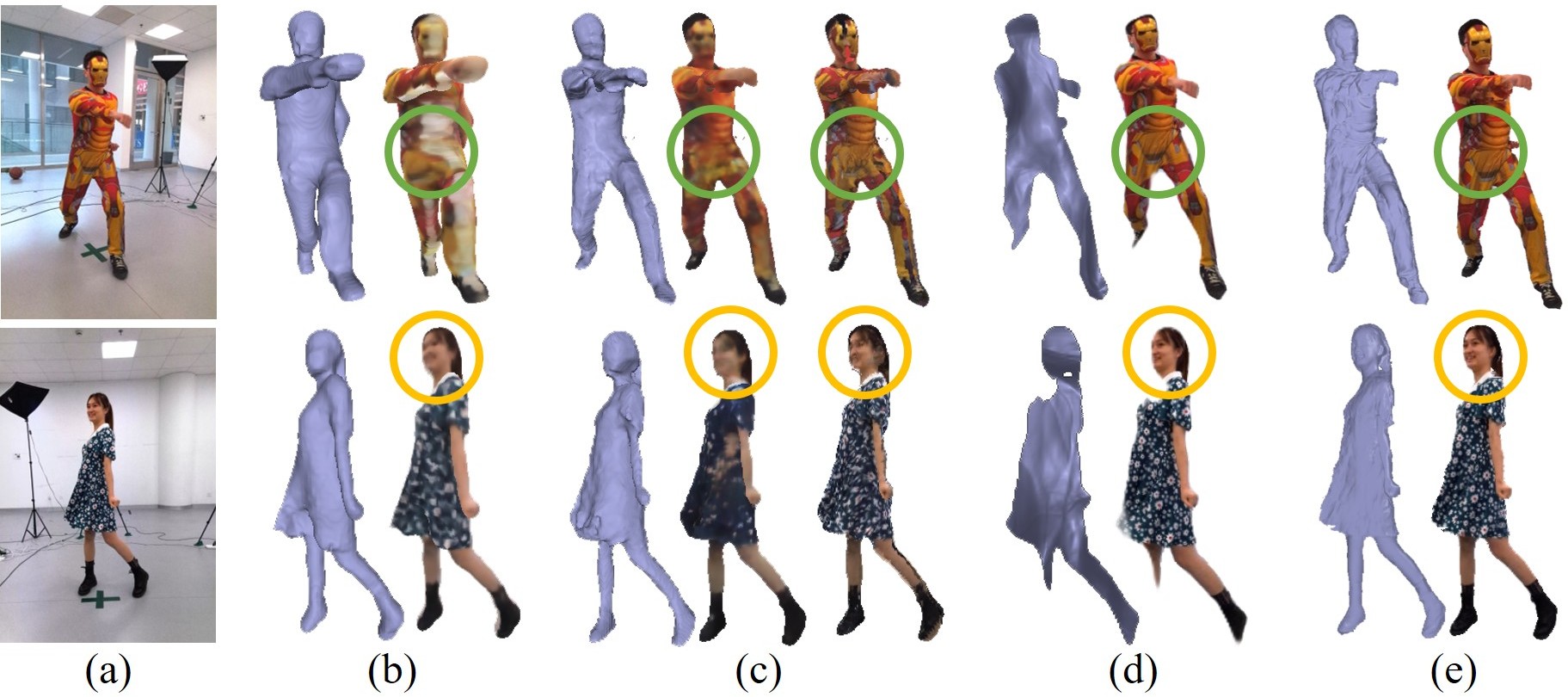} 
	\end{center} 
	\vspace{-10pt}
	\caption{Qualitative comparison. (a) Input images. (b-e) are the geometry and texture results from MonoPort~\cite{MonoPort}, Multi-PIFu~\cite{PIFU_2019ICCV}, Continuous View Control~\cite{CtViewControl} and ours, respectively. Note that the two texture results in (c) corresponds to the implicit texture and per-vertex texture, respectively.}
	\label{fig:fig_comp_1}
	\vspace{-10pt}
\end{figure*}

%Our testing dataset contains several dynamic sequences with six views, where 
In our real testing data, performers act complex motions with self-occlusion, and dress rich texture clothes, such as floral skirt and plaid shirt. 
We compare our NeuralHumanFVV against the state-of-the-art methods MonoPort~\cite{MonoPort}, Multi-PIFu~\cite{PIFU_2019ICCV} and Continuous View Control~\cite{CtViewControl} both in geometry and texture. 
%especially when the motions contain occupations and complex texture.

% ??????????
As shown in Fig.~\ref{fig:fig_comp_1}, our approach achieves significantly better texture results even when the texture is extraordinarily complex like the floral dress and the geometry we estimated filled with elegant details instead of a sketchy and bloated object. Even applying per-vertex texture mapping in the geometry from Multi-PIFu~\cite{PIFU_2019ICCV}, image blurs occur much more frequently in contrast to our results. 

\subsection{Comparison} 
\begin{table}[tpb]
	\footnotesize
	\newcommand{\tabincell}[2]{\begin{tabular}{@{}#1@{}}#2\end{tabular}}
	\newcolumntype{C}[1]{>{\centering\arraybackslash}p{#1}}
	\renewcommand{\arraystretch}{1.2}	
	\centering
	%\begin{tabular}{|C{1.6cm}|C{1.6cm}|C{1.6cm}|C{1.6cm}|}
	%	\hline\hline
	%	& {\scriptsize{RGB\_1}} &  \scriptsize{RGB\_6}  \\
	%	\hline
	%	\hline
	%	{\scriptsize{\textit{MonoPort}}}& 95.1  & 144.8  \\
	%	\hline
	%	{\scriptsize{\textit{Multi-PIFu}}}& 67.5  & 66.4  \\		
	%	\hline
	%	{\scriptsize{\textit{Multi-PIFu*}}}& 43.7  & 56.1  \\
	%	\hline
	%	{\scriptsize{\textit{CVC}}}& 98.1  & 103.1  \\
	%	\hline
	%	{\scriptsize{\textbf{Ours}}}& \textbf{27.6}  & \textbf{26.1}  \\			
	%	\hline
	%\end{tabular}
	%\begin{tabular}{cccccc}
	\begin{tabular}{C{0.8cm}C{0.9cm}C{0.9cm}C{0.9cm}C{1.0cm}C{1.4cm}}
	\hline\hline
    Method & \textit{MonoPort} & \textit{Multi-PIFu} & \textit{Multi-PIFu*} & \textit{CVC}   & \textbf{Ours} \\ \hline\hline
    RGB\_1 & 95.1$\pm$4.8     & 67.5$\pm$3.8       & 43.7$\pm$1.9        & 98.1$\pm$21.9  & \textbf{27.6} $\pm$ \textbf{1.6} \\\hline
    RGB\_6 & 144.8$\pm$6.6    & 66.4$\pm$2.2       & 56.1$\pm$2.1        & 103.1$\pm$27.4 & \textbf{26.1} $\pm$ \textbf{1.2}\\
    \hline
    \end{tabular}
    \vspace{4pt}
	\caption{Quantitative comparison of MonoPort~\cite{MonoPort}, Multi-PIFu~\cite{PIFU_2019ICCV}, Continuous View Control~\cite{CtViewControl} and NeuralHumanFVV. Multi-PIFu* denotes per-vertex texture mapping using the geometry from Multi-PIFu as input. RGB\_1 and RGB\_6 respectively present the MAE in one view and six views.  }
	\label{table:all} 
	\vspace{-10pt}
\end{table} 

Then, we make a quantitative comparison on the whole real testing dataset. We use mean absolute error (MAE) as the error metric. For overall MAE calculation, we average all MAEs from all images and frames.
%The RGB and depth value have been respectively normalized to $[0,255]$ and $[0,65525]$. 
Since Monoport~\cite{MonoPort} only takes one image as input, we also evaluate methods with single camera input (RGB\_1), compared with using all six cameras (RGB\_6).
As illustrated in Table.~\ref{table:all}, our approach surpasses other methods in all scenarios with distinct differences in overall MAE.

\begin{figure}[tbp] 
	\begin{center} 
		\includegraphics[width=0.9\linewidth]{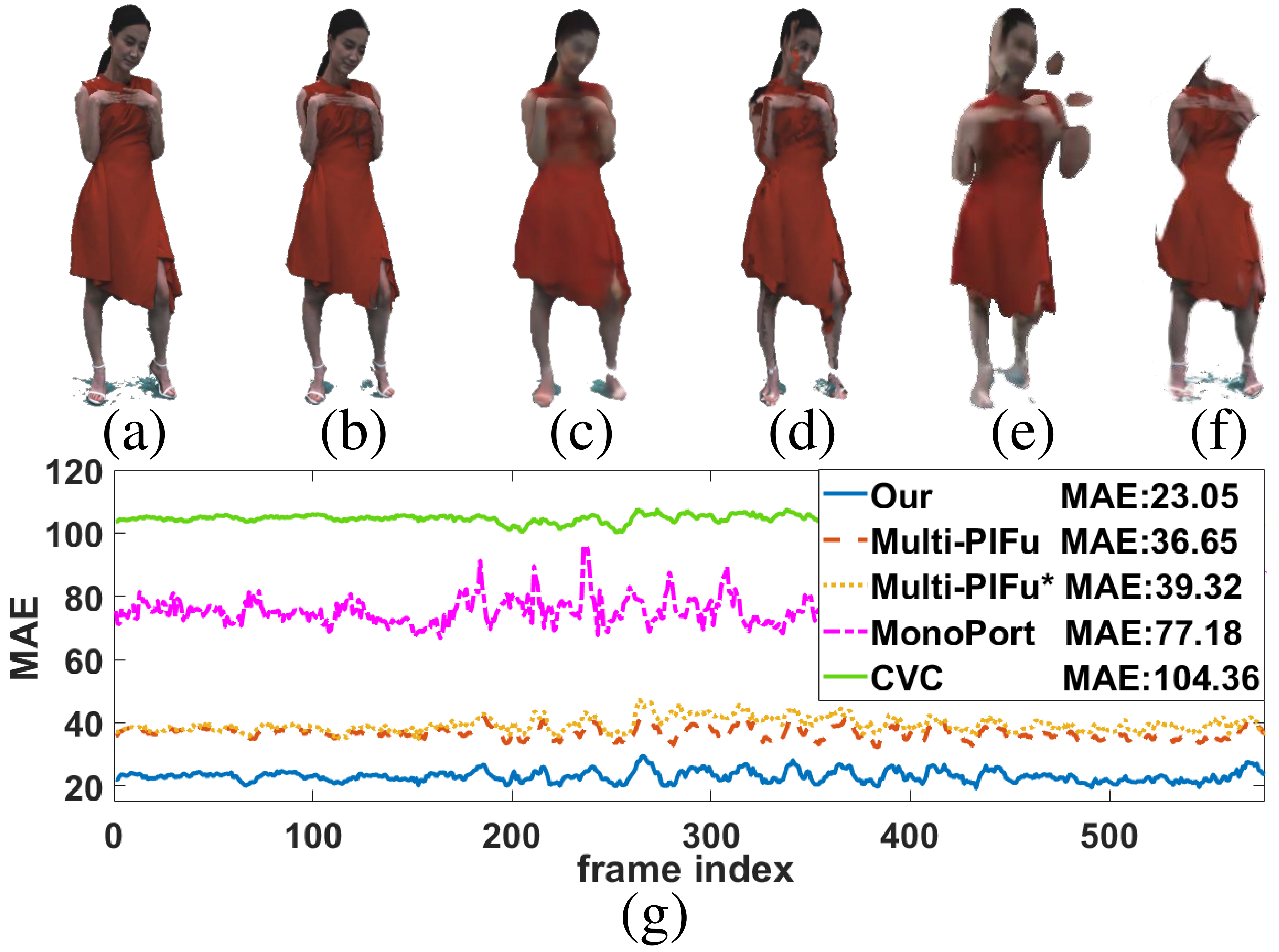} 
	\end{center} 
	\vspace{-10pt}
	\caption{Quantitative and qualitative comparison on synthesis sequence against NeuralHumanFVV, Multi-PIFu~\cite{PIFU_2019ICCV}, MonoPort~\cite{MonoPort} and Continuous View Control~\cite{CtViewControl}.(a) Color image in ground truth; (b) NeuralHumanFVV; (c) Multi-PIFu; (d)Multi-PIFu*  (e) MonoPort; (f)  Continuous View Control; (g)Error curves. Denote that Multi-PIFu* is the result of per-vertex texture mapping using the geometry from Multi-PIFu.} 
	\label{fig:fig_comp_2} 
	\vspace{-10pt}
\end{figure}

We make a comparison on a synthetic dynamic sequence with 600 frames, and generate 90 different target views to evaluate the MAE. The result is shown in Fig.~\ref{fig:fig_comp_2}. Our method can stay lowest MAE in the entire sequence.
%the advantage of our method becomes much distincter in the synthesis sequence testing on 90 different views in each frame as shown in Fig.~\ref{fig:fig_comp_2}. 

\subsection{Ablation Study} \label{sec:abla} 

\begin{figure}[tbp] 
	\begin{center} 
		\includegraphics[width=0.9\linewidth]{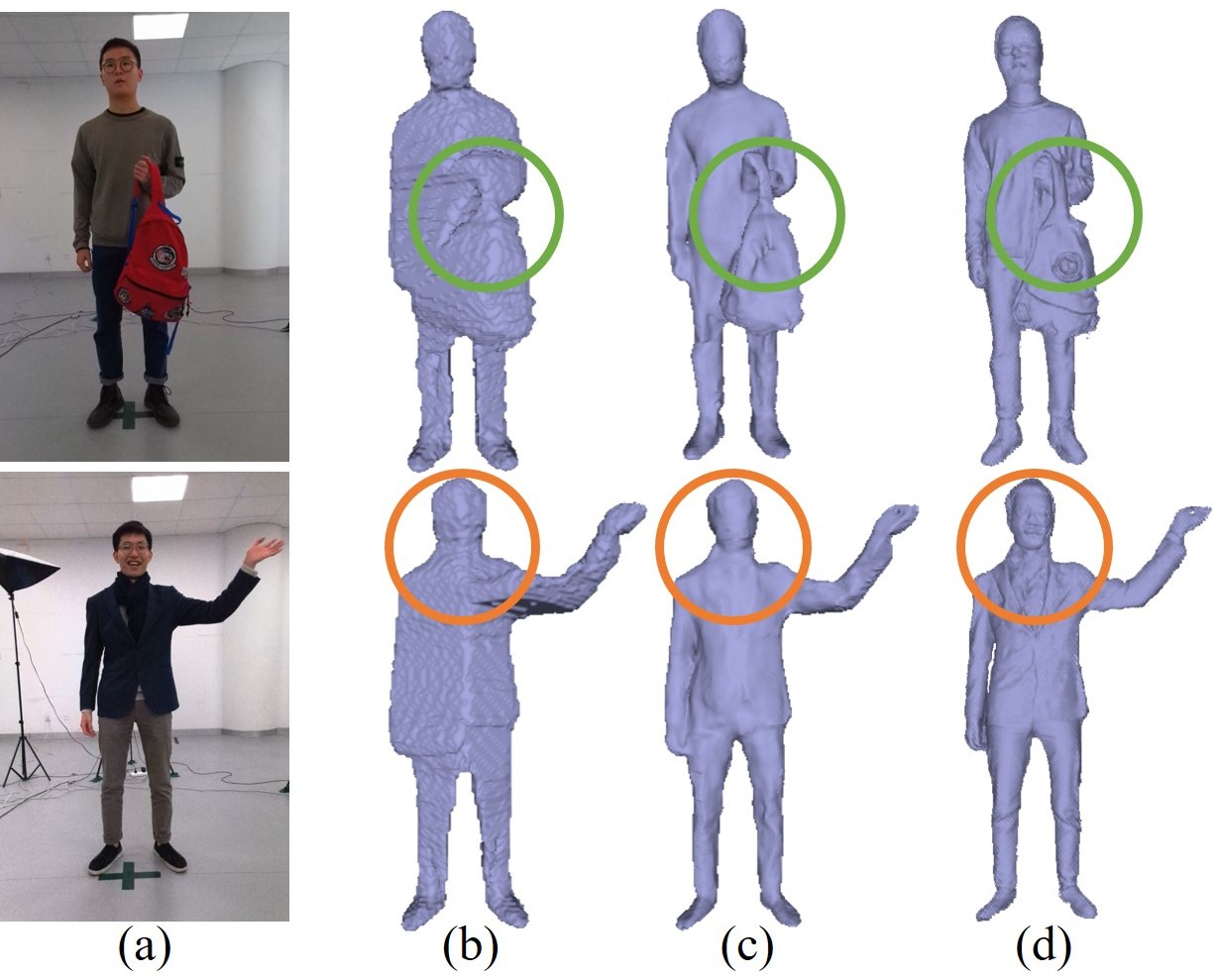} 
	\end{center} 
	\vspace{-10pt}
	\caption{Evaluation of our neural geometry generation. (a) Input images. (b) Geometry from SfS~\cite{sfs}; (c) Geometry without normal refinement; (d) Geometry from NeuralHumanFVV.} 
	\label{fig:fig_abla_geometry1} 
	\vspace{-5pt}
\end{figure}

\myparagraph{Neural Geometry Generation.}
Here, we evaluate our neural geometry generation scheme.
As shown in Fig.~\ref{fig:fig_abla_geometry1} (b), the results from SfS~\cite{sfs} only provide coarse geometry priors since only boundary information are utilized.
Our scheme without the normal refinement in Fig.~\ref{fig:fig_abla_geometry1} (c) can generate mid-level geometry details such as the clothing wrinkles but still suffers from over-smooth results, especially on the face regions.
In contrast, our approach with full pipeline in Fig.~\ref{fig:fig_abla_geometry1} (d) enables high-quality geometry detail generation almost with the level of details present in the input images.

%Per-vertex texture mapping applying the geometry from Multi-PIFu as input;

\begin{figure}[t] 
	\begin{center} 
		\includegraphics[width=0.95\linewidth]{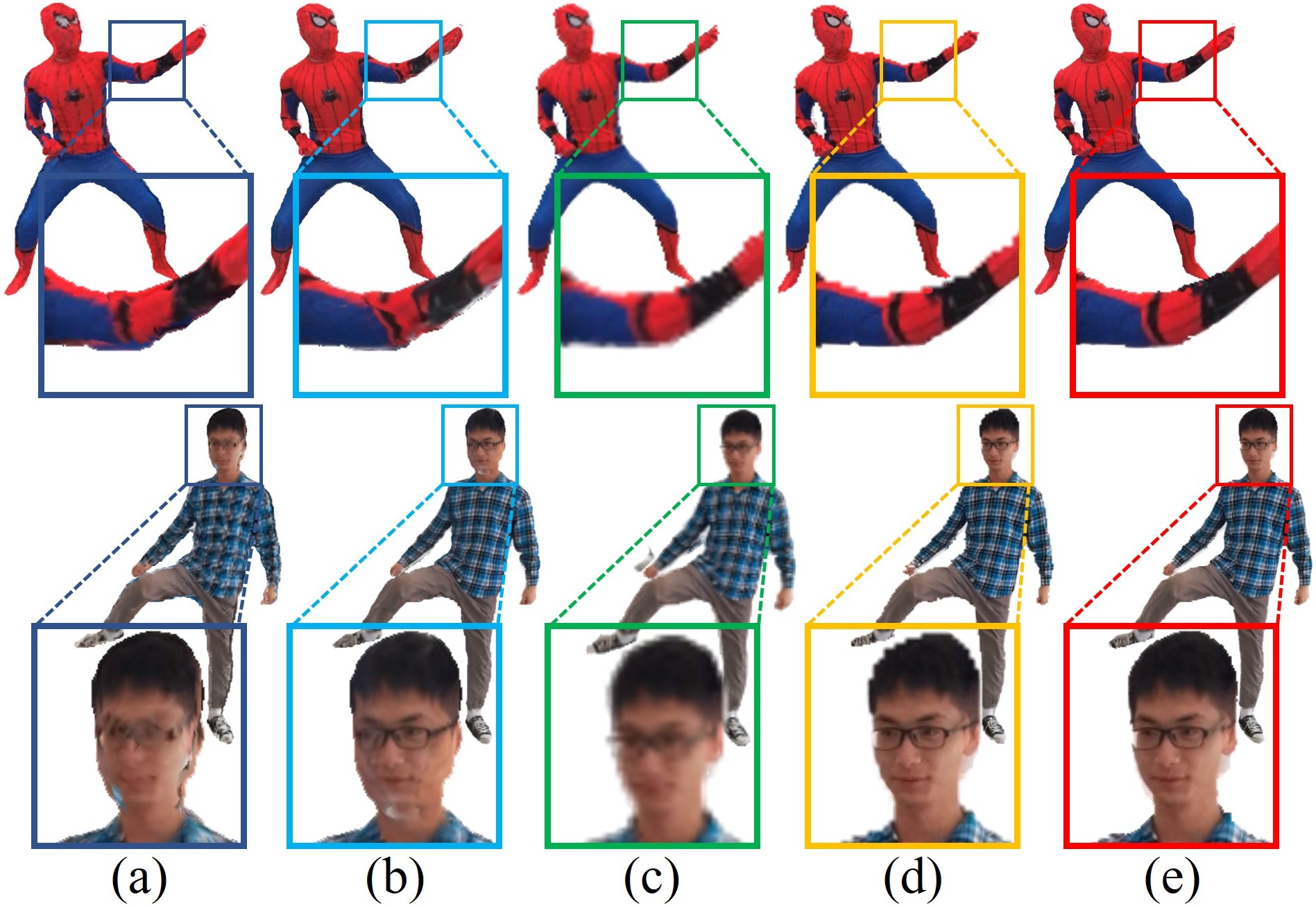} 
	\end{center} 
	\vspace{-10pt}
	\caption{Qualitative evaluation of our neural texture blending scheme. (a) Per-vertex texture mapping; (b) AGI\cite{AGISOFT}; (c) NeuralHumanFVV at $256\times256$ resolution ; (d) NeuralHumanFVV without boundary optimization; (e) NeuralHumanFVV.} 
	\label{fig:fig_abla_texture1.jpg} 
	\vspace{-10pt}
\end{figure}

\myparagraph{Neural Texture Blending.}
We further evaluate our neural texture blending scheme.
% Fig.9
In Fig.~\ref{fig:fig_abla_texture1.jpg}, we compare with our variations with different texturing schemes using the same geometry proxy. 
The per-vertex texturing in Fig.~\ref{fig:fig_abla_texture1.jpg} (a) suffers from severe block artifacts, while the offline scheme using the software AGI~\cite{AGISOFT} in Fig.~\ref{fig:fig_abla_texture1.jpg} (b) causes inferior results in those regions near the stitching seams.
%We also evaluate our neural scheme at low resolution situation in Fig.~\ref{fig:fig_abla_texture1.jpg} (c) and the one without the boundary optimization in Fig.~\ref{fig:fig_abla_texture1.jpg} (d), which suffer from over-smooth texture or coarse boundary, respectively. 
And our neural scheme at low resolution situation in Fig.~\ref{fig:fig_abla_texture1.jpg} (c) and the one without the boundary optimization in Fig.~\ref{fig:fig_abla_texture1.jpg} (d) suffer from over-smooth texture or coarse boundary, respectively. 
In contrast, our full neural texture scheme in Fig.~\ref{fig:fig_abla_texture1.jpg} (e) enables photo-realistic texture reconstruction in novel views.

% This not only highlights the contribution of each algorithmic component but also illustrates that our approach can recover the human texture with high accuracy and its robustness especially in challenging texture.

\begin{figure}[tbp] 
	\begin{center} 
		\includegraphics[width=0.9\linewidth]{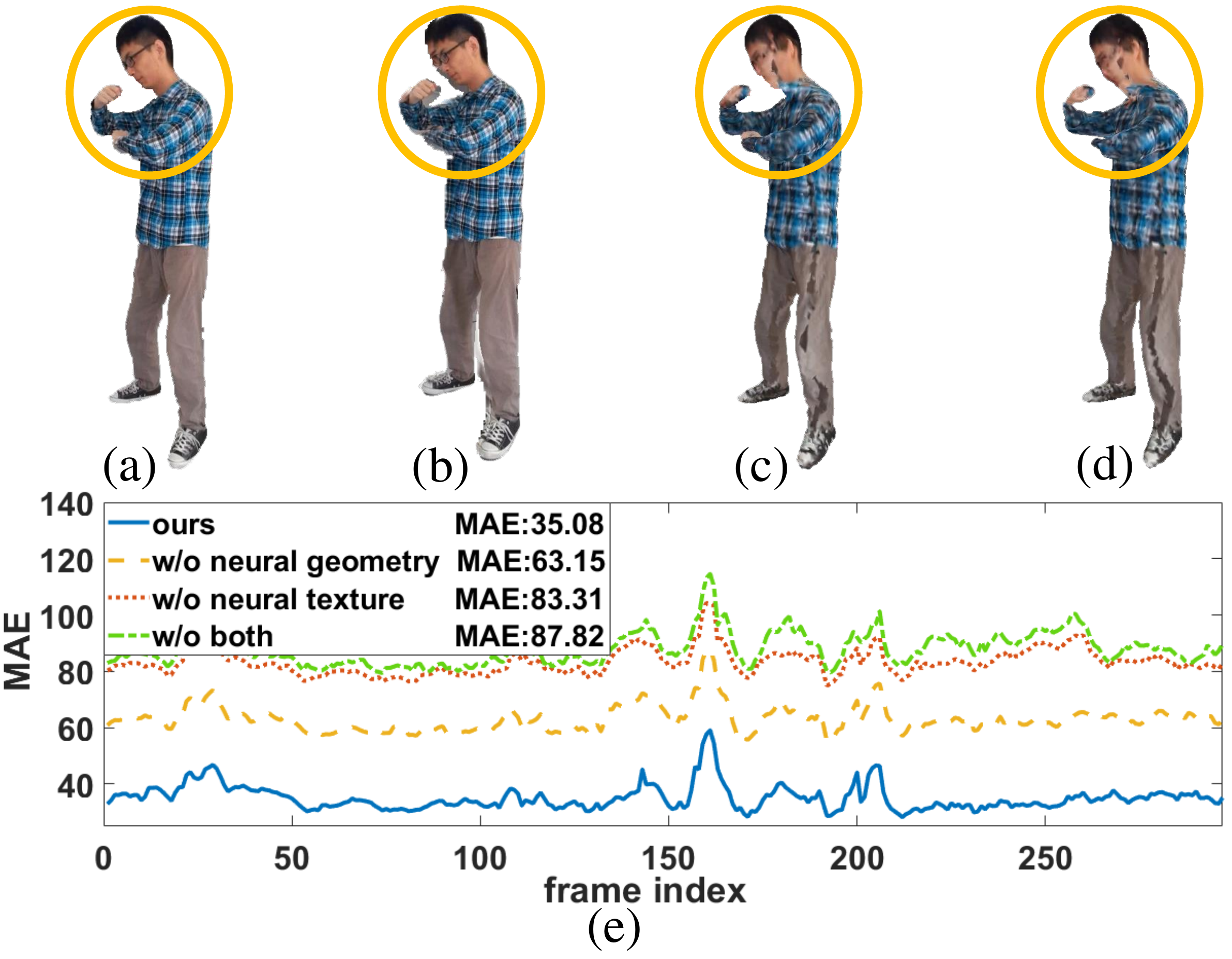} 
	\end{center} 
	\vspace{-10pt} 
	\caption{Quantitative evaluation. (a-d) The reconstructed results of ours, w/o neural geometry, w/o neural texture, w/o both.(e) Numerical error curves.} 
	\label{fig:fig_abla_texture2.jpg} 
	\vspace{-10pt}
\end{figure}

% fig.10 quantitative
For quantitative analysis of the individual components of NeuralHumanFVV, we utilize two different geometries as bases and two different texture methods to make a comparison among these four outputs as shown in as Fig.~\ref{fig:fig_abla_texture2.jpg}. Fig.~\ref{fig:fig_abla_texture2.jpg} (a) is from complete NeuralHumanFVV while Fig.~\ref{fig:fig_abla_texture2.jpg} (b) using the geometry from Multi-PIFu, Fig.~\ref{fig:fig_abla_texture2.jpg} (c) using the same geometry as Fig.~\ref{fig:fig_abla_texture2.jpg} (a) but per-vertex texture mapping, and Fig.~\ref{fig:fig_abla_texture2.jpg} (d) is yielded by Multi-PIFu and per-vertex texture mapping.
Not only our image outcome which has fewer artifacts but also the per-frame mean error for the three variation of our approach without model completion in Fig.~\ref{fig:fig_abla_texture2.jpg} (e) shows the advancement of our NeuralHumanFVV.

\begin{figure}[tbp] 
	\begin{center} 
		\includegraphics[width=0.9\linewidth]{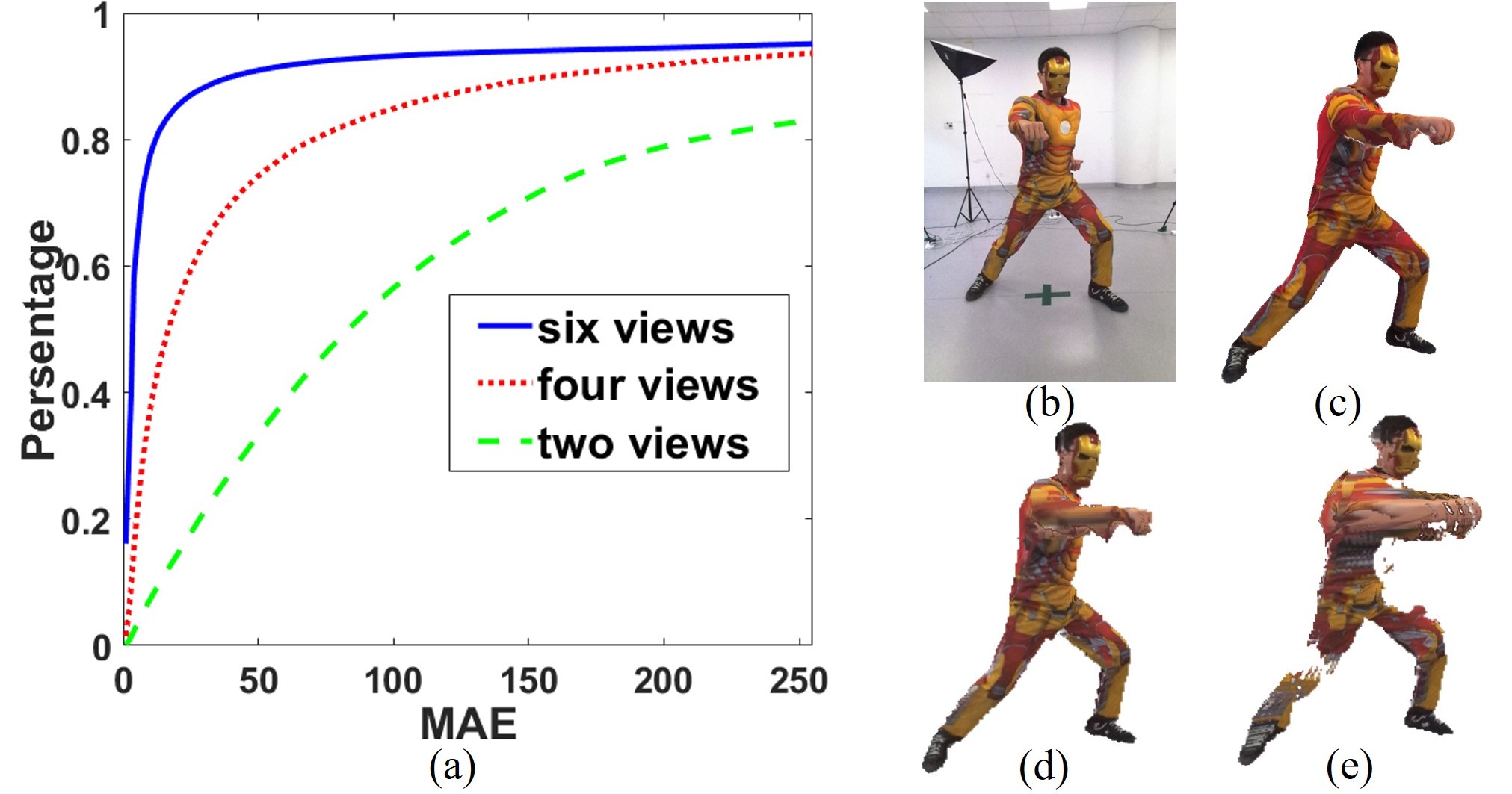} 
	\end{center} 
	\vspace{-10pt}
	\caption{Evaluation of the number input camera views. (a) Cumulative distribution function of the mean absolute error. (b) The reference capture scene. (c, d, e) Our reconstructed texture results using six, four and two cameras, respectively.} 
	\label{fig:fig_abla_camNum} 
	\vspace{-4mm}
\end{figure}

\myparagraph{Camera Number.}
To evaluate the influence of input views in our multi-view setting, we compare to the variation of our pipeline using various numbers of input camera views.
As shown in Fig.~\ref{fig:fig_abla_camNum}, the reconstruction results without enough camera views suffer from severe geometry and blending artifacts and the average error increases significantly as the the camera number decreases.
Empirically, the setting with six cameras serve as a good compromising settlement.

\section {Discussion} 

\myparagraph{Limitation.}
As the first trial to enable real-time and photo-realistic neural human performance capture and rendering from only sparse RGB inputs, the proposed NeuralHumanFVV system still owns some limitations.
First, inaccuracy of segmentation leads to incomplete regions in our final synthesized images.
Our texturing results depend on the input image resolutions. Thin structures like fingers are difficult to reconstruct due to the input with limited resolution.
Our system generates plausible geometry detail from RGB images but physically inaccurate when the testing images deviate much from the training ones.

\myparagraph{Conclusion.}We have presented a real-time neural performance rendering system to generate high-quality geometry and photo-realistic textures of human activities in novel views only using sparse multiple RGB cameras.
Our neural geometry generation benefits inherently from our multi-view setting and enables efficient and implicit reasoning of underlying geometry in a novel view.
Our neural blending scheme with occlusion analysis and boundary-aware upsampling further enables to recover high resolution (e.g., 1k) and photo-realistic textures without sacrificing the real-time performance.
Our experimental results demonstrate the effectiveness of NeuralHumanFVV for high-quality human performance rendering in challenging scenarios with various poses, clothing types and topology changes.
% , which compares favorably to the state-of-the-arts.
We believe that our approach is a critical step to virtually but realistic teleport human performances, with many potential applications in VR/AR like gaming, entertainment and immersive telepresense.

{\small
\bibliographystyle{ieee_fullname}
\bibliography{NeuralHumanFVV_arxiv}

\begin{thebibliography}{10}\itemsep=-1pt

\bibitem{twindom}
Twindom dataset.
\newblock \url{https://https://web.twindom.com/}.

\bibitem{AGISOFT}
Agisoft photoscan professional.
\newblock http://www.agisoft.com/downloads/installer/, 2019.

\bibitem{aliev2019neural}
Kara-Ali Aliev, Artem Sevastopolsky, Maria Kolos, Dmitry Ulyanov, and Victor
  Lempitsky.
\newblock Neural point-based graphics.
\newblock {\em arXiv preprint arXiv:1906.08240}, 2019.

\bibitem{TEX2SHAPE_2019ICCV}
Thiemo Alldieck, Gerard Pons-Moll, Christian Theobalt, and Marcus Magnor.
\newblock Tex2shape: Detailed full human body geometry from a single image.
\newblock In {\em The IEEE International Conference on Computer Vision (ICCV)},
  October 2019.

\bibitem{SCAPE2005}
Dragomir Anguelov, Praveen Srinivasan, Daphne Koller, Sebastian Thrun, Jim
  Rodgers, and James Davis.
\newblock Scape: Shape completion and animation of people.
\newblock In {\em ACM SIGGRAPH 2005 Papers}, SIGGRAPH ’05, page 408–416,
  New York, NY, USA, 2005. Association for Computing Machinery.

\bibitem{bhatnagar2019mgn}
Bharat~Lal Bhatnagar, Garvita Tiwari, Christian Theobalt, and Gerard Pons-Moll.
\newblock Multi-garment net: Learning to dress 3d people from images.
\newblock In {\em {IEEE} International Conference on Computer Vision ({ICCV})}.
  {IEEE}, oct 2019.

\bibitem{BreglM1998}
Chris Bregler and Jitendra Malik.
\newblock Tracking people with twists and exponential maps.
\newblock In {\em Computer Vision and Pattern Recognition (CVPR)}, 1998.

\bibitem{OpenPose}
Zhe Cao, Tomas Simon, Shih-En Wei, and Yaser Sheikh.
\newblock Realtime multi-person 2d pose estimation using part affinity fields.
\newblock In {\em Computer Vision and Pattern Recognition (CVPR)}, 2017.

\bibitem{CtViewControl}
Xu Chen, Jie Song, and Otmar Hilliges.
\newblock Monocular neural image based rendering with continuous view control.
\newblock In {\em Proceedings of the IEEE/CVF International Conference on
  Computer Vision (ICCV)}, pages 4089--4099, October 2019.

\bibitem{sfs}
Kong~Man Cheung, Simon Baker, and Takeo Kanade.
\newblock Shape-from-silhouette of articulated objects and its use for human
  body kinematics estimation and motion capture.
\newblock In {\em 2003 IEEE Computer Society Conference on Computer Vision and
  Pattern Recognition, 2003. Proceedings.}, volume~1, pages I--I, 2003.

\bibitem{collet2015high}
Alvaro Collet, Ming Chuang, Pat Sweeney, Don Gillett, Dennis Evseev, David
  Calabrese, Hugues Hoppe, Adam Kirk, and Steve Sullivan.
\newblock High-quality streamable free-viewpoint video.
\newblock {\em ACM Transactions on Graphics (TOG)}, 34(4):69, 2015.

\bibitem{Curless1996}
Brian Curless and Marc Levoy.
\newblock A volumetric method for building complex models from range images.
\newblock In {\em Proceedings of the 23rd Annual Conference on Computer
  Graphics and Interactive Techniques}, SIGGRAPH '96, pages 303--312, New York,
  NY, USA, 1996. ACM.

\bibitem{motion2fusion}
Mingsong Dou, Philip Davidson, Sean~Ryan Fanello, Sameh Khamis, Adarsh Kowdle,
  Christoph Rhemann, Vladimir Tankovich, and Shahram Izadi.
\newblock Motion2fusion: Real-time volumetric performance capture.
\newblock {\em ACM Trans. Graph.}, 36(6):246:1--246:16, Nov. 2017.

\bibitem{dou-siggraph2016}
Mingsong Dou, Sameh Khamis, Yury Degtyarev, Philip Davidson, Sean Fanello,
  Adarsh Kowdle, Sergio~Orts Escolano, Christoph Rhemann, David Kim, Jonathan
  Taylor, Pushmeet Kohli, Vladimir Tankovich, and Shahram Izadi.
\newblock {Fusion4D: Real-time Performance Capture of Challenging Scenes}.
\newblock In {\em ACM SIGGRAPH Conference on Computer Graphics and Interactive
  Techniques}, 2016.

\bibitem{Ganapathi10}
Varun Ganapathi, Christian Plagemann, Daphne Koller, and Sebastian Thrun.
\newblock Real time motion capture using a single time-of-flight camera.
\newblock 2010.

\bibitem{TheRelightables}
Kaiwen Guo, Peter Lincoln, Philip Davidson, Jay Busch, Xueming Yu, Matt Whalen,
  Geoff Harvey, Sergio Orts-Escolano, Rohit Pandey, Jason Dourgarian, and et
  al.
\newblock The relightables: Volumetric performance capture of humans with
  realistic relighting.
\newblock {\em ACM Trans. Graph.}, 38(6), Nov. 2019.

\bibitem{guo2017real}
Kaiwen Guo, Feng Xu, Tao Yu, Xiaoyang Liu, Qionghai Dai, and Yebin Liu.
\newblock Real-time geometry, albedo and motion reconstruction using a single
  rgbd camera.
\newblock {\em ACM Transactions on Graphics (TOG)}, 2017.

\bibitem{DensePose}
Rıza~Alp Güler, Natalia Neverova, and Iasonas Kokkinos.
\newblock Densepose: Dense human pose estimation in the wild.
\newblock In {\em Proceedings of the IEEE Conference on Computer Vision and
  Pattern Recognition (CVPR)}, June 2018.

\bibitem{LiveCap2019tog}
Marc Habermann, Weipeng Xu, Michael Zollh\"{o}fer, Gerard Pons-Moll, and
  Christian Theobalt.
\newblock Livecap: Real-time human performance capture from monocular video.
\newblock {\em ACM Transactions on Graphics (TOG)}, 38(2):14:1--14:17, 2019.

\bibitem{DeepBlending}
Peter Hedman, Julien Philip, True Price, Jan-Michael Frahm, George Drettakis,
  and Gabriel Brostow.
\newblock Deep blending for free-viewpoint image-based rendering.
\newblock {\em ACM Trans. Graph.}, 37(6), Dec. 2018.

\bibitem{DeepVolumetric_2018ECCV}
Zeng Huang, Tianye Li, Weikai Chen, Yajie Zhao, Jun Xing, Chloe LeGendre,
  Linjie Luo, Chongyang Ma, and Hao Li.
\newblock Deep volumetric video from very sparse multi-view performance
  capture.
\newblock In {\em The European Conference on Computer Vision (ECCV)}, September
  2018.

\bibitem{DodgeABullet_ECCV2018}
Shi Jin, Ruiynag Liu, Yu Ji, Jinwei Ye, and Jingyi Yu.
\newblock Learning to dodge a bullet: Concyclic view morphing via deep
  learning.
\newblock In Vittorio Ferrari, Martial Hebert, Cristian Sminchisescu, and Yair
  Weiss, editors, {\em Computer Vision -- ECCV 2018}, pages 230--246, Cham,
  2018. Springer International Publishing.

\bibitem{joo2015panoptic}
Hanbyul Joo, Hao Liu, Lei Tan, Lin Gui, Bart Nabbe, Iain Matthews, Takeo
  Kanade, Shohei Nobuhara, and Yaser Sheikh.
\newblock {Panoptic Studio: A Massively Multiview System for Social Motion
  Capture}.
\newblock In {\em Proceedings of the IEEE International Conference on Computer
  Vision}, pages 3334--3342, 2015.

\bibitem{TotalCapture}
Hanbyul Joo, Tomas Simon, and Yaser Sheikh.
\newblock Total capture: A 3d deformation model for tracking faces, hands, and
  bodies.
\newblock In {\em The IEEE Conference on Computer Vision and Pattern
  Recognition (CVPR)}, June 2018.

\bibitem{HMR18}
Angjoo Kanazawa, Michael~J. Black, David~W. Jacobs, and Jitendra Malik.
\newblock End-to-end recovery of human shape and pose.
\newblock In {\em Computer Vision and Pattern Regognition (CVPR)}, 2018.

\bibitem{RTH_nvs}
Youngjoong Kwon, Stefano Petrangeli, Dahun Kim, Haoliang Wang, Eunbyung Park,
  Viswanathan Swaminathan, and Henry Fuchs.
\newblock Rotationally-temporally consistent novel view synthesis of human
  performance video.
\newblock In Andrea Vedaldi, Horst Bischof, Thomas Brox, and Jan-Michael Frahm,
  editors, {\em Computer Vision -- ECCV 2020}, pages 387--402, Cham, 2020.
  Springer International Publishing.

\bibitem{lazova3dv2019}
Verica Lazova, Eldar Insafutdinov, and Gerard Pons-Moll.
\newblock 360-degree textures of people in clothing from a single image.
\newblock In {\em International Conference on 3D Vision (3DV)}, sep 2019.

\bibitem{li2009robust}
Hao Li, Bart Adams, Leonidas~J Guibas, and Mark Pauly.
\newblock Robust single-view geometry and motion reconstruction.
\newblock 28(5):175, 2009.

\bibitem{HaoliTemplate}
Hao Li, Linjie Luo, Daniel Vlasic, Pieter Peers, Jovan Popovi\'{c}, Mark Pauly,
  and Szymon Rusinkiewicz.
\newblock Temporally coherent completion of dynamic shapes.
\newblock In {\em ACM Trans. Graph.}, volume~31, Feb. 2012.

\bibitem{MonoPort}
Ruilong Li, Yuliang Xiu, Shunsuke Saito, Zeng Huang, Kyle Olszewski, and Hao
  Li.
\newblock Monocular real-time volumetric performance capture.
\newblock In Andrea Vedaldi, Horst Bischof, Thomas Brox, and Jan-Michael Frahm,
  editors, {\em Computer Vision -- ECCV 2020}, pages 49--67, Cham, 2020.
  Springer International Publishing.

\bibitem{liu2020neural}
Lingjie Liu, Jiatao Gu, Kyaw~Zaw Lin, Tat-Seng Chua, and Christian Theobalt.
\newblock Neural sparse voxel fields.
\newblock {\em NeurIPS}, 2020.

\bibitem{liu2013markerless}
Yebin Liu, Juergen Gall, Carsten Stoll, Qionghai Dai, Hans-Peter Seidel, and
  Christian Theobalt.
\newblock Markerless motion capture of multiple characters using multiview
  image segmentation.
\newblock {\em Pattern Analysis and Machine Intelligence, IEEE Transactions
  on}, 35(11):2720--2735, 2013.

\bibitem{NeuralVolumes}
Stephen Lombardi, Tomas Simon, Jason Saragih, Gabriel Schwartz, Andreas
  Lehrmann, and Yaser Sheikh.
\newblock Neural volumes: Learning dynamic renderable volumes from images.
\newblock {\em ACM Trans. Graph.}, 38(4), July 2019.

\bibitem{SMPL2015}
Matthew Loper, Naureen Mahmood, Javier Romero, Gerard Pons-Moll, and Michael~J.
  Black.
\newblock Smpl: A skinned multi-person linear model.
\newblock {\em ACM Trans. Graph.}, 34(6):248:1--248:16, Oct. 2015.

\bibitem{LookinGood}
Ricardo Martin-Brualla, Rohit Pandey, Shuoran Yang, Pavel Pidlypenskyi,
  Jonathan Taylor, Julien Valentin, Sameh Khamis, Philip Davidson, Anastasia
  Tkach, Peter Lincoln, and et al.
\newblock Lookingood: Enhancing performance capture with real-time neural
  re-rendering.
\newblock {\em ACM Trans. Graph.}, 37(6), Dec. 2018.

\bibitem{Mehta2017}
Dushyant Mehta, Srinath Sridhar, Oleksandr Sotnychenko, Helge Rhodin, Mohammad
  Shafiei, Hans-Peter Seidel, Weipeng Xu, Dan Casas, and Christian Theobalt.
\newblock Vnect: Real-time 3d human pose estimation with a single rgb camera.
\newblock {\em ACM Transactions on Graphics (TOG)}, 36(4), 2017.

\bibitem{occupancy2019CVPR}
Lars Mescheder, Michael Oechsle, Michael Niemeyer, Sebastian Nowozin, and
  Andreas Geiger.
\newblock Occupancy networks: Learning 3d reconstruction in function space.
\newblock In {\em The IEEE Conference on Computer Vision and Pattern
  Recognition (CVPR)}, June 2019.

\bibitem{NRWird_CVPR19}
Moustafa Meshry, Dan~B. Goldman, Sameh Khamis, Hugues Hoppe, Rohit Pandey, Noah
  Snavely, and Ricardo Martin-Brualla.
\newblock Neural rerendering in the wild.
\newblock In {\em Proceedings of the IEEE/CVF Conference on Computer Vision and
  Pattern Recognition (CVPR)}, June 2019.

\bibitem{nerf}
Ben Mildenhall, Pratul~P. Srinivasan, Matthew Tancik, Jonathan~T. Barron, Ravi
  Ramamoorthi, and Ren Ng.
\newblock Nerf: Representing scenes as neural radiance fields for view
  synthesis.
\newblock In Andrea Vedaldi, Horst Bischof, Thomas Brox, and Jan-Michael Frahm,
  editors, {\em Computer Vision -- ECCV 2020}, pages 405--421, Cham, 2020.
  Springer International Publishing.

\bibitem{Newcombe2015}
Richard~A. Newcombe, Dieter Fox, and Steven~M. Seitz.
\newblock {DynamicFusion: Reconstruction and Tracking of Non-Rigid Scenes in
  Real-Time}.
\newblock June 2015.

\bibitem{KinectFusion}
Richard~A. Newcombe, Shahram Izadi, Otmar Hilliges, David Molyneaux, David Kim,
  Andrew~J. Davison, Pushmeet Kohli, Jamie Shotton, Steve Hodges, and Andrew
  Fitzgibbon.
\newblock {KinectFusion: Real-Time Dense Surface Mapping and Tracking}.
\newblock In {\em Proc. of ISMAR}, pages 127--136, 2011.

\bibitem{SemiPara_2019CVPR}
Rohit Pandey, Anastasia Tkach, Shuoran Yang, Pavel Pidlypenskyi, Jonathan
  Taylor, Ricardo Martin-Brualla, Andrea Tagliasacchi, George Papandreou,
  Philip Davidson, Cem Keskin, Shahram Izadi, and Sean Fanello.
\newblock Volumetric capture of humans with a single rgbd camera via
  semi-parametric learning.
\newblock In {\em The IEEE Conference on Computer Vision and Pattern
  Recognition (CVPR)}, June 2019.

\bibitem{DeepSDF}
Jeong~Joon Park, Peter Florence, Julian Straub, Richard Newcombe, and Steven
  Lovegrove.
\newblock Deepsdf: Learning continuous signed distance functions for shape
  representation.
\newblock In {\em Proceedings of the IEEE/CVF Conference on Computer Vision and
  Pattern Recognition (CVPR)}, June 2019.

\bibitem{SMPLX2019}
Georgios Pavlakos, Vasileios Choutas, Nima Ghorbani, Timo Bolkart, Ahmed A.~A.
  Osman, Dimitrios Tzionas, and Michael~J. Black.
\newblock Expressive body capture: 3d hands, face, and body from a single
  image.
\newblock In {\em Proceedings IEEE Conf. on Computer Vision and Pattern
  Recognition (CVPR)}, pages 10975--10985, June 2019.

\bibitem{People3D_2019ICCV}
Albert Pumarola, Jordi Sanchez-Riera, Gary P.~T. Choi, Alberto Sanfeliu, and
  Francesc Moreno-Noguer.
\newblock 3dpeople: Modeling the geometry of dressed humans.
\newblock In {\em The IEEE International Conference on Computer Vision (ICCV)},
  October 2019.

\bibitem{FreeViewSynthesis}
Gernot Riegler and Vladlen Koltun.
\newblock Free view synthesis.
\newblock In Andrea Vedaldi, Horst Bischof, Thomas Brox, and Jan-Michael Frahm,
  editors, {\em Computer Vision -- ECCV 2020}, Cham, 2020. Springer
  International Publishing.

\bibitem{ronneberger2015u}
Olaf Ronneberger, Philipp Fischer, and Thomas Brox.
\newblock U-net: Convolutional networks for biomedical image segmentation.
\newblock In {\em International Conference on Medical image computing and
  computer-assisted intervention}, pages 234--241. Springer, 2015.

\bibitem{PIFU_2019ICCV}
Shunsuke Saito, Zeng Huang, Ryota Natsume, Shigeo Morishima, Angjoo Kanazawa,
  and Hao Li.
\newblock Pifu: Pixel-aligned implicit function for high-resolution clothed
  human digitization.
\newblock In {\em The IEEE International Conference on Computer Vision (ICCV)},
  October 2019.

\bibitem{PIFuHD}
Shunsuke Saito, Tomas Simon, Jason Saragih, and Hanbyul Joo.
\newblock Pifuhd: Multi-level pixel-aligned implicit function for
  high-resolution 3d human digitization.
\newblock In {\em Proceedings of the IEEE/CVF Conference on Computer Vision and
  Pattern Recognition (CVPR)}, June 2020.

\bibitem{Shotton2011}
J. Shotton, A. Fitzgibbon, M. Cook, T. Sharp, M. Finocchio, R. Moore, A.
  Kipman, and A. Blake.
\newblock {Real-time Human Pose Recognition in Parts from Single Depth Images}.
\newblock 2011.

\bibitem{DeepVoxels_CVPR2019}
Vincent Sitzmann, Justus Thies, Felix Heide, Matthias Niessner, Gordon
  Wetzstein, and Michael Zollhofer.
\newblock Deepvoxels: Learning persistent 3d feature embeddings.
\newblock In {\em Proceedings of the IEEE/CVF Conference on Computer Vision and
  Pattern Recognition (CVPR)}, June 2019.

\bibitem{SRN_nips2019}
Vincent Sitzmann, Michael Zollhoefer, and Gordon Wetzstein.
\newblock Scene representation networks: Continuous 3d-structure-aware neural
  scene representations.
\newblock In H. Wallach, H. Larochelle, A. Beygelzimer, F. d\textquotesingle
  Alch\'{e}-Buc, E. Fox, and R. Garnett, editors, {\em Advances in Neural
  Information Processing Systems}, volume~32, pages 1121--1132. Curran
  Associates, Inc., 2019.

\bibitem{KillingFusion2017cvpr}
M. Slavcheva, M. Baust, D. Cremers, and S. Ilic.
\newblock {KillingFusion: Non-rigid 3D Reconstruction without Correspondences}.
\newblock In {\em IEEE Conference on Computer Vision and Pattern Recognition
  (CVPR)}, 2017.

\bibitem{StollHGST2011}
Carsten Stoll, Nils Hasler, Juergen Gall, Hans-Peter Seidel, and Christian
  Theobalt.
\newblock Fast articulated motion tracking using a sums of {Gaussians} body
  model.
\newblock In {\em International Conference on Computer Vision (ICCV)}, 2011.

\bibitem{robustfusion}
Zhuo Su, Lan Xu, Zerong Zheng, Tao Yu, Yebin Liu, and Lu Fang.
\newblock Robustfusion: Human volumetric capture with data-driven visual cues
  using a rgbd camera.
\newblock In Andrea Vedaldi, Horst Bischof, Thomas Brox, and Jan-Michael Frahm,
  editors, {\em Computer Vision -- ECCV 2020}, pages 246--264, Cham, 2020.
  Springer International Publishing.

\bibitem{sumner2007embedded}
Robert~W Sumner, Johannes Schmid, and Mark Pauly.
\newblock Embedded deformation for shape manipulation.
\newblock {\em ACM Transactions on Graphics (TOG)}, 26(3):80, 2007.

\bibitem{DetailDepth_2019ICCV}
Sicong Tang, Feitong Tan, Kelvin Cheng, Zhaoyang Li, Siyu Zhu, and Ping Tan.
\newblock A neural network for detailed human depth estimation from a single
  image.
\newblock In {\em The IEEE International Conference on Computer Vision (ICCV)},
  October 2019.

\bibitem{Templaterealtime}
J. {Taylor}, J. {Shotton}, T. {Sharp}, and A. {Fitzgibbon}.
\newblock The vitruvian manifold: Inferring dense correspondences for one-shot
  human pose estimation.
\newblock In {\em 2012 IEEE Conference on Computer Vision and Pattern
  Recognition}, pages 103--110, 2012.

\bibitem{NR_survey}
Ayush Tewari, Ohad Fried, Justus Thies, Vincent Sitzmann, Stephen Lombardi,
  Kalyan Sunkavalli, Ricardo Martin-Brualla, Tomas Simon, Jason Saragih,
  Matthias Nießner, Rohit Pandey, Sean Fanello, Gordon Wetzstein, Jun-Yan Zhu,
  Christian Theobalt, Maneesh Agrawala, Eli Shechtman, Dan~B. Goldman, and
  Michael Zollhöfer.
\newblock {State of the Art on Neural Rendering}.
\newblock {\em Computer Graphics Forum}, 2020.

\bibitem{TheobASST2010}
Christian Theobalt, Edilson de Aguiar, Carsten Stoll, Hans-Peter Seidel, and
  Sebastian Thrun.
\newblock Performance capture from multi-view video.
\newblock In {\em Image and Geometry Processing for 3-D Cinematography}, pages
  127--149. Springer, 2010.

\bibitem{Thies2020Image-guided}
Justus Thies, Michael Zollhöfer, Christian Theobalt, Marc Stamminger, and
  Matthias Nießner.
\newblock Image-guided neural object rendering.
\newblock In {\em International Conference on Learning Representations}, 2020.

\bibitem{Wu2013}
Chenglei Wu, Carsten Stoll, Levi Valgaerts, and Christian Theobalt.
\newblock On-set performance capture of multiple actors with a stereo camera.
\newblock 32(6), 2013.

\bibitem{Wu_2020_CVPR}
Minye Wu, Yuehao Wang, Qiang Hu, and Jingyi Yu.
\newblock Multi-view neural human rendering.
\newblock In {\em Proceedings of the IEEE/CVF Conference on Computer Vision and
  Pattern Recognition (CVPR)}, June 2020.

\bibitem{Xiang_2019_CVPR}
Donglai Xiang, Hanbyul Joo, and Yaser Sheikh.
\newblock Monocular total capture: Posing face, body, and hands in the wild.
\newblock In {\em The IEEE Conference on Computer Vision and Pattern
  Recognition (CVPR)}, June 2019.

\bibitem{FlyFusion}
L. {Xu}, W. {Cheng}, K. {Guo}, L. {Han}, Y. {Liu}, and L. {Fang}.
\newblock Flyfusion: Realtime dynamic scene reconstruction using a flying depth
  camera.
\newblock {\em IEEE Transactions on Visualization and Computer Graphics}, pages
  1--1, 2019.

\bibitem{FlyCap}
Lan Xu, Yebin Liu, Wei Cheng, Kaiwen Guo, Guyue Zhou, Qionghai Dai, and Lu
  Fang.
\newblock Flycap: Markerless motion capture using multiple autonomous flying
  cameras.
\newblock {\em IEEE Transactions on Visualization and Computer Graphics},
  24(8):2284--2297, Aug 2018.

\bibitem{UnstructureLan}
L. {Xu}, Z. {Su}, L. {Han}, T. {Yu}, Y. {Liu}, and L. {FANG}.
\newblock Unstructuredfusion: Realtime 4d geometry and texture reconstruction
  using commercialrgbd cameras.
\newblock {\em IEEE Transactions on Pattern Analysis and Machine Intelligence},
  pages 1--1, 2019.

\bibitem{EventCap}
Lan Xu, Weipeng Xu, Vladislav Golyanik, Marc Habermann, Lu Fang, and Christian
  Theobalt.
\newblock Eventcap: Monocular 3d capture of high-speed human motions using an
  event camera.
\newblock In {\em Proceedings of the IEEE/CVF Conference on Computer Vision and
  Pattern Recognition (CVPR)}, June 2020.

\bibitem{MonoPerfCap}
Weipeng Xu, Avishek Chatterjee, Michael Zollh\"{o}fer, Helge Rhodin, Dushyant
  Mehta, Hans-Peter Seidel, and Christian Theobalt.
\newblock Monoperfcap: Human performance capture from monocular video.
\newblock {\em ACM Transactions on Graphics (TOG)}, 37(2):27:1--27:15, 2018.

\bibitem{DVS_photo}
Zexiang Xu, Sai Bi, Kalyan Sunkavalli, Sunil Hadap, Hao Su, and Ravi
  Ramamoorthi.
\newblock Deep view synthesis from sparse photometric images.
\newblock {\em ACM Trans. Graph.}, 38(4), July 2019.

\bibitem{DoubleFusion}
Tao Yu, Zerong Zheng, Kaiwen Guo, Jianhui Zhao, Qionghai Dai, Hao Li, Gerard
  Pons-Moll, and Yebin Liu.
\newblock Doublefusion: Real-time capture of human performances with inner body
  shapes from a single depth sensor.
\newblock {\em Transactions on Pattern Analysis and Machine Intelligence
  (TPAMI)}, 2019.

\bibitem{DeepHuman_2019ICCV}
Zerong Zheng, Tao Yu, Yixuan Wei, Qionghai Dai, and Yebin Liu.
\newblock Deephuman: 3d human reconstruction from a single image.
\newblock In {\em The IEEE International Conference on Computer Vision (ICCV)},
  October 2019.

\bibitem{zollhofer2014real}
Michael Zollh{\"o}fer, Matthias Nie{\ss}ner, Shahram Izadi, Christoph Rehmann,
  Christopher Zach, Matthew Fisher, Chenglei Wu, Andrew Fitzgibbon, Charles
  Loop, Christian Theobalt, et~al.
\newblock {Real-time Non-rigid Reconstruction using an RGB-D Camera}.
\newblock {\em ACM Transactions on Graphics (TOG)}, 33(4):156, 2014.

\end{thebibliography}
}

\end{document}